\documentclass[lettersize, journal]{IEEEtran}
\usepackage{amsmath,amsfonts}
\usepackage{algorithmic}
\usepackage{algorithm}
\usepackage{array}
\usepackage[caption=false,font=normalsize,labelfont=sf,textfont=sf]{subfig}
\usepackage{cite}
\usepackage{color}
\definecolor{light_gray}{gray}{.9}
\usepackage{colortbl}  %彩色表格需要加载的宏包
\usepackage{xcolor}

\usepackage[pagebackref=false,breaklinks=false,letterpaper=true,colorlinks,citecolor=blue,linkcolor=blue, anchorcolor=blue, bookmarks=false]{hyperref}
\usepackage{booktabs}
\usepackage{multirow}
\definecolor{highlight}{RGB}{255, 255, 204}

\usepackage{bbding}
\newcommand{\cmark}{\Checkmark}%
\newcommand{\xmark}{\XSolidBrush}%

\usepackage{textcomp}
\usepackage{stfloats}
\usepackage{url}
\usepackage{verbatim}
\usepackage{graphicx}
\usepackage{cite}
\usepackage{lettrine} % startpr
\newcommand{\et}{\emph{et al. }}

\usepackage{tikz}
\usepackage[edges]{forest}
\usetikzlibrary{arrows.meta}

\begin{document}

\title{When Geoscience Meets Foundation Models: Towards General Geoscience Artificial Intelligence System}

\author{Hao Zhang,~\IEEEmembership{Graduate Student Member,~IEEE,} Jin-Jian Xu, Hong-Wei Cui, Lin Li, \\
Yaowen Yang, Chao-Sheng Tang, Niklas Boers
        % <-this % stops a space
\thanks{
  
This work was supported by the China Scholarship Council (Grant No. 202306830108, 202206190069), the Postgraduate Research \& Practice Innovation Program of Jiangsu Province (Grant No. KYCX23\_0380), the Interdisciplinary Innovation Fund for Doctoral Students of NUAA (No. KXKCXJJ202302), the National Natural Science Foundation of China (Grant No. 41925012, 42230710), Natural Science Foundation of Jiangsu Province (Grant No. BK20211087), and the National Research Foundation, Singapore under its AI Singapore Programme (AISG Award No: AISG2-TC-2021-001). This research is also supported by NB acknowledges funding by the Volkswagen Foundation. (\emph{Correspondence: Jin-Jian Xu, Chao-Sheng Tang, and Niklas Boers}.)

Hao Zhang is with the College of Electronic and Information Engineering, Nanjing University of Aeronautics and Astronautics, Nanjing 211106 China. He is also with the Key Laboratory of Dynamic Cognitive System of Electromagnetic Spectrum Space (Nanjing University of Aeronautics and Astronautics) and with the Ministry of Industry and Information Technology, Nanjing, 211106, China. He is also with School of Electrical and Electronic Engineering, Nanyang Technological University, 50 Nanyang Avenue, Singapore 639798. (email: haozhangcn@nuaa.edu.cn).

Jin-Jian Xu, Lin Li, and Chao-Sheng Tang are with the School of Earth Sciences and Engineering, Nanjing University, 163 Xianlin Road, Nanjing 210023, China. (email: xujinjian@nju.edu.cn, linli@smail.nju.edu.cn, tangchaosheng@nju.edu.cn)

Hong-Wei Cui is with Huawei Technologies Co., Ltd, Shanghai 200040, China. (email:cuihongwei@pku.edu.cn)

Yaowen Yang is with the School of Civil and Environmental Engineering, Nanyang Technological University, Blk N1, 50 7 Nanyang Ave, Singapore 639798, Singapore. (email: cywyang@ntu.edu.sg)

Niklas Boers is with the Earth System Modelling, School of Engineering and Design, Technical University of Munich, Munich, Germany. He is also with the Potsdam Institute for Climate Impact Research, Telegrafenberg A31, 14473 Potsdam, Germany. (email: n.boers@tum.de)

}
}

\markboth{Accepted by IEEE Geoscience and Remote Sensing Magazine}%
{Zhang \MakeLowercase{\textit{et al.}}: When Geoscience Meets Foundation Models: Towards General Geoscience Artificial Intelligence System}


\maketitle

\begin{abstract}
Artificial intelligence (AI) has significantly advanced Earth sciences, yet its full potential in to comprehensively modeling Earth's complex dynamics remains unrealized. Geoscience foundation models (GFMs) emerge as a paradigm-shifting solution, integrating extensive cross-disciplinary data to enhance the simulation and understanding of Earth system dynamics. These data-centric AI models extract insights from petabytes of structured and unstructured data, effectively addressing the complexities of Earth systems that traditional models struggle to capture. The unique strengths of GFMs include flexible task specification, diverse input-output capabilities, and multi-modal knowledge representation, enabling analyses that surpass those of individual data sources or traditional AI methods. This review not only highlights the key advantages of GFMs, but also presents essential techniques for their construction, with a focus on transformers, pre-training, and adaptation strategies. Subsequently, we examine recent advancements in GFMs, including large language models, vision models, vision-language models, and foundation-model-based agents, particularly emphasizing the potential applications in remote sensing. Additionally, the review concludes with a comprehensive analysis of the challenges and future trends in GFMs, addressing five critical aspects: data integration, model complexity, uncertainty quantification, interdisciplinary collaboration, and concerns related to privacy, trust, and security. This review offers a comprehensive overview of emerging geoscientific research paradigms, emphasizing the untapped opportunities at the intersection of advanced AI techniques and geoscience. It examines major methodologies, showcases advances in large-scale models, and discusses the challenges and prospects that will shape the future landscape of GFMs. The paper highlights a dynamic field rich with possibilities, poised to unlock new insights into Earth's complexities and further advance geoscience exploration.
\end{abstract}


\section{Introduction}
\lettrine[lines=2]{A}{S} Earth science enters the era of big data, artificial intelligence (AI) holds significant potential not only for addressing classic geoscientific problems and enhancing understanding of Earth's complex, interactive, and multiscale processes \cite{bergen2019machine,li2023big}. The availability of massive volumes of Earth system data, already exceeding dozens of petabytes with hundreds of terabytes transmitted daily, has facilitated the widespread adoption of AI technologies, in data-driven Earth system sciences \cite{reichstein2019deep,deng2024geoscience}. The utilization of AI not only enhances geological events (\emph{e.g.}, earthquakes and volcanic eruptions) through real-time seismic data analysis but also holds significant value in bias correction and upsampling of Earth system models \cite{anantrasirichai2019deep,bergen2019machine,bischoff2023unpaired,hess2022physically,deng2023learning}. Machine learning approaches support the exploration and mapping of natural resources (\emph{e.g.}, minerals and oil), thereby increasing accuracy and minimizing the ecological impacts of mining and drilling operations \cite{ham2019deep,mitsui2021seasonal,li2022intelligent}. Moreover, the application of deep learning algorithms in climate modeling and weather forecasting plays a crucial role in understanding the effects of climate change, refining weather predictions, and enhancing disaster management and preparedness \cite{weyn2019can,ravuri2021skilful,lam2022graphcast,bi2023accurate,chen2023fengwu,zhang2023skilful}. Additionally, the advanced capabilities of AI in processing and interpreting extensive amounts of satellite imagery and sensor data also makes a substantial contribution to environmental monitoring and conservation, which enables a diverse array of critical applications, including environmental monitoring, land cover classification, urban planning, disaster response, and climate change analysis \cite{sun2022ringmo,wang2022advancing,jakubik2023foundation,ding2024adapting}.

The successful applications of AI technologies in earth science present formidable challenges and inherent limitations that necessitate careful consideration. A predominant issue is the dependency on extensive high-quality datasets, which are frequently scarce or incomplete in geoscience disciplines stemming from the complex and variable nature of Earth systems \cite{karpatne2018machine,arrowsmith2022big}. Furthermore, the generalization capability of AI models is frequently compromised when transitioning between different geological contexts. This dependence on specific training data can result in significant inaccuracies when models are applied to novel or poorly understood environments \cite{briscoe2023mechanistic,aslam2024wetland,chen2024rsmamba}. Another critical limitation is the interpretability of AI outputs. For AI-driven insights to be effectively incorporated into geoscience research and practical applications, they must be transparent and comprehensible, thereby fostering trust and adoption among geoscientists \cite{toms2020pyhsically,shen2023differentiable,jiang2024interpretable}. To address these challenges, the development of larger, more robust models is imperative for advancing data-driven Earth system science. 

Recently, foundation models have emerged as pivotal tools for tackling many complex data-centric tasks through pre-training on vast datasets and subsequent fine-tuning for specific applications \cite{bommasani2021opportunities}. Foundation models exhibit unparalleled emergent capabilities, including contextual reasoning, complex problem-solving, and zero-shot adaptability across diverse tasks, which are driven by advanced algorithms, extensive training on vast datasets, and sophisticated neural network architectures \cite{vartinen2022generating,hu2023promptcap,van2023chatgpt,yan2023ringmo}. The generalization capability of foundation models distinctly sets them apart from traditional AI models, which were confined to specific, singular tasks and exhibited limited performance on diverse tasks beyond their original scope. In contrast, foundation models undergo extensive pre-training on expansive, varied datasets, enabling them to develop comprehensive representations of language and visual data \cite{nie2024survey}. For instance, as of 2024, the large language models (LLMs), such as ChatGPT-4 \cite{achiam2023gpt} and Claude-3/3.5 \cite{claude3haiku,claude3-5}, have integrated capabilities in vision, speech, and knowledge graphs, impressing users with their ability to produce text indistinguishable from human writing and generate images and computer code from diverse prompts \cite{chen2023fengwu,peng2023instruction,van2023chatgpt,xie2024wildfiregpt}. Moreover, many contemporary foundation models are capable of processing both images and text, finding applications in diverse fields such as medicine, transportation, remote sensing, and urban planning \cite{steinberg2021language,yang2022large,nori2023capabilities,singhal2023large,li2024vision,zhang2024earthgpt}. Given the transformative potential of such technologies, it is imperative to develop specialized foundation models for geoscience, which can revolutionize the field by providing deep, actionable insights into complex Earth systems.

Although there have been preliminary efforts to develop geoscience foundation models (GFMs) \cite{lacoste2021toward,fuller2022satvit,deng2023learning,hong2024spectralgpt,li2024vision,manvi2024large,zhang2024earthgpt}, widespread adoption within geoscience AI remains limited. The slow progression in GFMs is attributed to persistent challenges in accessing and processing multi-source and multi-modal geoscience datasets, the inherent complexity of Earth systems, and the inherent limitations of current AI technologies \cite{cui2023generalized,kasneci2023chatgpt,li2024interpretable}. Although the task-specific approach persists as the dominant paradigm in geoscience AI. These innovations in GFMs may transform how geoscientists process and interpret data, leading to enhanced accuracy and impact in critical research areas such as sustainability, climate change, extreme weather events, natural hazards, and environmental monitoring \cite{sahil2023development,zhang2024towards,zhu2024foundations}. Therefore, the adoption of adaptable, multifaceted foundation models in place of narrow, static task-specific models has the potential to not only provide new insights and capabilities to Earth science, but also to transform our understanding of complex Earth systems, drive innovation in interdisciplinary research, and address global challenges with greater precision and foresight.

This study methodically traces the evolution of AI applications within the earth sciences, highlighting the limitations of traditional AI methodologies. It underscores the growing importance of foundation models in geosciences, analyzes the transformative impact of these models on geoscience research paradigms, and explores potential research directions for GFMs. The study also elaborates on the unique capabilities of GFMs, such as their adaptability to diverse tasks, their proficiency in processing multimodal data, and their seamless integration of complex geoscientific knowledge. Together, these characteristics demonstrate the potential of GFMs to redefine task specification, enhance multimodal capabilities, and strengthen geoscientific knowledge representations, thereby establishing a new paradigm and standard for geoscience AI.

\subsection{Existing Surveys and Reviews}

Table \ref{tab:survey} provides an overview of existing surveys and reviews related to foundation models for geoscience. These works have explored various aspects of foundation models in geoscience, including remote sensing, urban science, vision-language models, generative artificial intelligence, geophysical foundation models, and geospatial location embedding. 
These works are published in 2023 and 2024. For example, 
Jiao \et \cite{jiao2023brain} contributed to the field with their proposed brain-inspired framework for remote sensing foundation models. 
In a Chinese-language publication, Zhang's team \cite{zhang2023large} offered a comprehensive review of pre-training techniques for large remote sensing models. 
The landscape of smart cities is enriched by the work's \cite{zhang2024towards} introduction of Urban Foundation Models. 
The authors in \cite{li2024vision} delved into the intersection of vision-language models and remote sensing in their examination. 
The potential of generative AI and large language models in geoscience is explored in depth by the research group led by Hadid \et \cite{hadid2024geoscience}. 
A thought-provoking discussion on geophysical foundation models in exploration geophysics was presented in the perspective piece by the work \cite{liu2024foundation}. 
Zhou \et spearheaded a thorough review \cite{zhou2024towards} that sheds light on vision-language geo-foundation models. 
The authors in \cite{tucker2024systematic} break new ground with their investigation into geospatial location embedding in large language models.
However, there is still a need for a comprehensive survey that integrates these diverse perspectives and provides a holistic view of the current state of the field. 
This paper aims to fill this gap by providing a comprehensive review of the key advantages, applications, techniques, and advances in GFMs. We also discuss the challenges and future trends of GFMs in geoscience, highlighting the potential of these models to revolutionize Earth system science.

\begin{table*}[htbp]
  \scriptsize % font size
  \centering
  \caption{Existing Surveys and Reviews Related to Foundation Models for Geoscience}
    \begin{tabular}{m{0.5cm}<{\centering}m{2cm}<{\centering}m{11cm}<{\centering}m{0.5cm}<{\centering}m{0.5cm}<{\centering}m{0.5cm}<{\centering}}
    \toprule
    Year &  Publication & Summary & Advs. & Techs. & Adv. \\
    \midrule
    2023 & Jiao \et \cite{jiao2023brain} & This survey examines foundation models in remote sensing, comparing their performance across various tasks and datasets, and proposes a brain-inspired framework for remote sensing foundation models while identifying open research problems.   & \cmark & \xmark & \cmark \\
    \midrule
    2023 & Zhang \et \cite{zhang2023large} & This review examines pre-training techniques for large remote sensing models, analyzing supervised and unsupervised single-modal models as well as visual-text joint multimodal models, and discusses future prospects for integrating domain knowledge, enhancing generalization, expanding applications, and reducing data costs. (in Chinese)   & \cmark & \xmark &  \xmark \\    
    \midrule
    2024 & Zhang \et \cite{zhang2024towards} & This paper introduces urban foundation models (UFMs), proposes a data-centric taxonomy for current UFM-related works, presents a framework for their future development, and explores their potential applications in smart cities.   &\cmark  & \xmark &\cmark  \\
    \midrule
    2024 & Li \et \cite{li2024vision} & This review comprehensively examines the application of vision-language models (VLMs) in remote sensing, covering various tasks such as image captioning, text-based image generation and retrieval, visual question answering, and scene analysis, while discussing current progress, challenges, and future research directions in this emerging field.  &\cmark  & \xmark &\cmark  \\
    \midrule
    2024 & Hadid \et \cite{hadid2024geoscience} & This survey explores the applications and potential of generative artificial intelligence (GAI) and large language models in geoscience, discussing various GAI models used in the field, their applications, challenges, and future promises in addressing Earth system dynamics and related scientific domains.    &\cmark  & \xmark &\cmark  \\
    \midrule
    2024 & Liu \et \cite{liu2024foundation} &  This perspective discusses the potential applications, research directions, and challenges of developing geophysical foundation models (GeoFMs) in exploration geophysics and related sub-disciplines, proposing strategies for their construction and exploring new research paradigms in the field. &\cmark  & \xmark &\cmark  \\
    \midrule
    2024 & Zhou \et \cite{zhou2024towards} & This paper provides a comprehensive review of vision-language geo-foundation models (VLGFMs), covering their background, core technologies, applications, and future research directions in the field of multimodal geospatial data analysis.   &\cmark  & \xmark &\cmark  \\
    \midrule
    2024 & Tucker \et \cite{tucker2024systematic} & This review explores geospatial location embedding (GLE) in large language models, identifying four main themes and proposing a spatial foundation/language mdel (SLM) framework to advance spatial artificial intelligence systems (SPAIS) with embedded spatial knowledge.    & \cmark & \xmark &  \cmark\\
    
    \midrule
    \rowcolor{highlight}\textbf{2024}  & \textbf{This work} & \textbf{A comprehensive review of geoscience foundation models (GFMs) from the view of key advantages, key applications, key techniques, and key advances}. &  \cmark & \cmark & \cmark \\
    \bottomrule
    \end{tabular}
  \label{tab:survey}
\end{table*}

\subsection{Contributions and Organization}

Though there exists a growing body of literature on foundation models in geoscience, the field is still in its infancy, with many open research questions and challenges. Table \ref{tab:survey} provides an overview of existing surveys and reviews related to foundation models for geoscience. These works have explored various aspects of foundation models in geoscience, including remote sensing, urban science, vision-language models, generative artificial intelligence, geophysical foundation models, and geospatial location embedding. However, there is still a need for a comprehensive survey that integrates these diverse perspectives and provides a holistic view of the current state of the field. This paper aims to fill this gap by providing a comprehensive review of the key advantages, techniques, and advances in geoscience foundation models. We also discuss the challenges and future trends of GFMs in geoscience, highlighting the potential of these models to revolutionize Earth system science. 
The main contributions of this paper are as follows
\begin{itemize}
  \item This review provides a comprehensive overview of geoscience foundation models (GFMs) from the perspective of key advantages, key applications, key techniques, and key advances. The study also discusses the challenges and future trends of GFMs in geoscience, emphasizing the potential of these models to revolutionize Earth system science.
  \item This study introduces a new paradigm by integrating AI foundation models with specialized geoscience knowledge. It addresses the unique challenges of geoscience data, including its spatial-temporal complexity and interdisciplinary nature. The research aims to develop a general geoscience AI system capable of handling diverse tasks across space, air, ground, and ocean domains. This integration enhances the scalability and generalizability of AI models in geoscience, potentially transforming how researchers approach complex earth science problems.
  \item We present a comprehensive review of the key advances in GFMs, including large language models (LLMs), large vision models (LVMs), large vision-language models (LVLMs), and foundation-model-based agents, with a particular emphasis on their applications in remote sensing. We also discuss the potential of GFMs to revolutionize Earth system science by providing deep, actionable insights into complex Earth systems. 
  \item This study contributes a thorough analysis of challenges and future trends in GFMs. We identify key obstacles including data complexity, model interpretability, uncertainty quantification, interdisciplinary collaboration, and privacy, trust   and security concerns. This forward-looking perspective guides strategic development and innovation in the rapidly evolving field of geoscience AI.
\end{itemize}

\begin{figure*}[!t]
\centering
\includegraphics[width=0.95\textwidth]{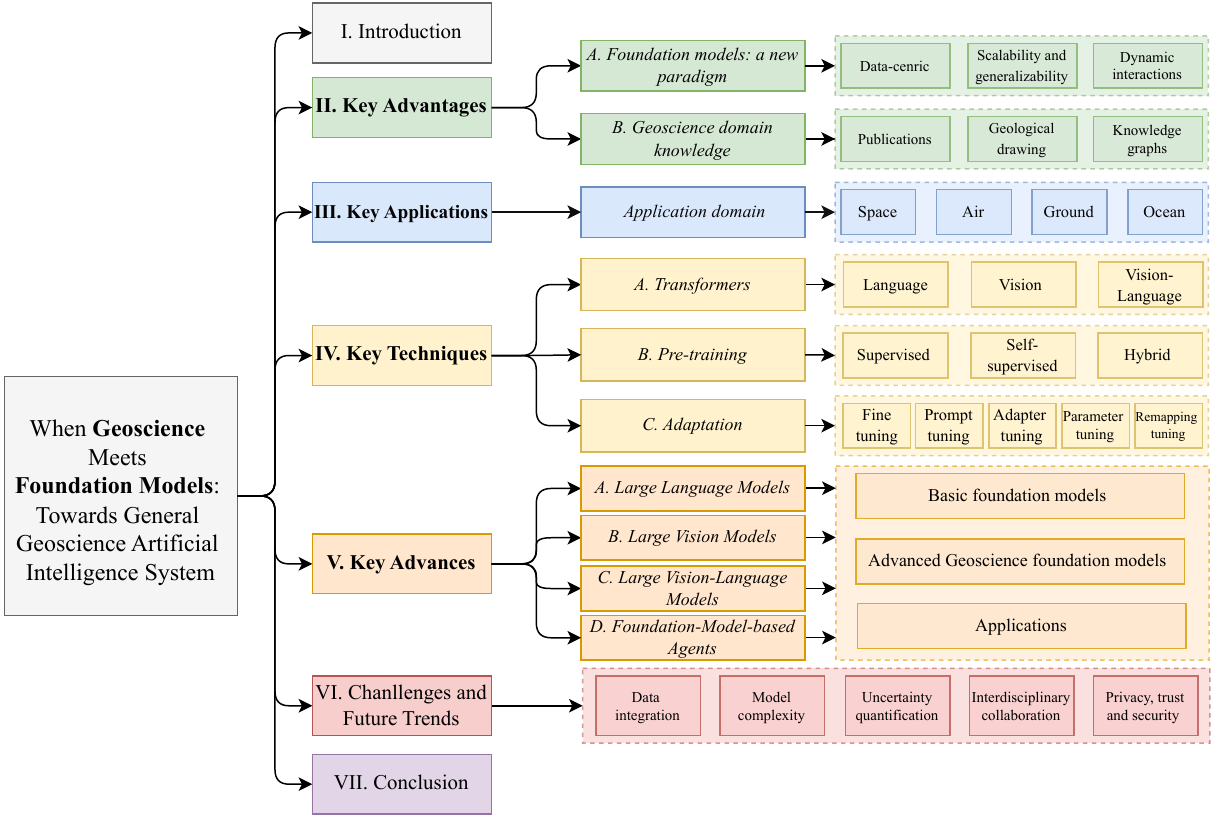}
\caption{Organization of the review.}
\label{fig:overview}
\end{figure*}

The remainder of this paper is organized as follows. 
Section \ref{sec:key_advantages} discusses the key advantages of GFMs, including their data-centric paradigm, scalability, generalizability, dynamic interactions, and causal inference. 
Section \ref{sec:key_applications} explores the potential applications of GFMs in Earth science, including space, air, ground, and ocean domains. 
Section \ref{sec:key_techniques} introduces the geoscience domain knowledge and the potential applications of GFMs in Earth science. Section \ref{sec:key_advances} presents the recent advances in GFMs, including large language models, large vision models, large vision-language models, and foundation-model-based agents, especially remote sensing applications. Section \ref{sec:challenges_and_future_trends} discusses the challenges and future trends of GFMs. Finally, Section \ref{sec:conclusion} concludes the paper.

\section{Key Advantages of Geoscience Foundation Models}
\label{sec:key_advantages}

In this section, we discuss the key advantages of geoscience foundation models (GFMs) from the view of a new paradigm, Geoscience domain knowledge, and application domains. Data-centric AI, capability and generalizability, and dynamic interactions make GFMs a new paradigm compared to traditional AI. Geoscience domain knowledge including publications, geological data, and knowledge graph further enhance the capabilities of GFMs. The application domain of GFMs in Earth science are also discussed, including space, air, ground, and ocean domains. 

\subsection{Foundation Models: A New Paradigm}
Machine learning and deep learning have been widely applied for solving problems in Earth system science \cite{reichstein2019deep,bergen2019machine,irrgang2021towards,li2024new}. Machine learning focuses on “how” to define the features of data and make decisions using learning algorithms, while deep learning infers “features” automatically by designing specific architectures. Foundation models focus on functionalities such as in-context learning by using large pre-trained models. They rely on massive, diverse data, large pre-trained models, and self-supervised training, yet exhibit even larger computing demands.

\subsubsection{Data-Centric AI}
Traditional machine learning methods focus on developing models and algorithms for predefined datasets and in particular deep learning involves unsupervised learning tasks. However, real-world applications involve working with noisy, imperfect data, and uncertain labels. Rather than treating datasets as fixed, performance can be substantially improved by refining the data itself. This has given rise to the emerging field of data-centric AI (DCAI), which aims to systematically engineer better datasets as a means to improve model performance. Whereas skilled data scientists have long manually curated datasets in an ad hoc fashion based on intuition, DCAI seeks to establish more formal and objective principles and techniques for dataset optimization. This represents a paradigm shift from perceiving data as a static input to models, to co-developing datasets and models in an integrated fashion. Further research in DCAI may uncover new methods for improved data collection, automated data cleaning, labeling, augmentation, and representation learning that make the overall machine learning process more robust and efficient.

Drawing inspiration from foundational models beyond geoscience, we have identified three distinct features that set GFMs apart from traditional geoscience machine and deep learning models. Firstly, employing GFMs for novel tasks will be as straightforward as articulating the tasks in natural language processing, such as English or other languages. GFMs will possess the capability to address previously unencountered challenges through dynamic task specification, enabling them to comprehend and tackle new tasks without the need for retraining. Secondly, GFMs can accept inputs and generate outputs utilizing various combinations of data modalities. This flexibility allows for the incorporation of different types of data, such as videos, images, sound, text, laboratory results, or any combination thereof, into the model. This flexible interactivity capability contrasts with the constraints of more stringent traditional multimodal models. Thirdly, GFMs will employ a formal representation of geoscientific knowledge, enabling it to engage in reasoning processes for previously unseen tasks and provide interpretations using geoscientifically accurate language.

\subsubsection{Scalability and Generalizability} % (fold)
\label{ssub:scalability_and_generalizability}
Foundation models include large amounts of data, large models, requiring large computing capacity, self-training, all enabling wide ranges of applications, exhibiting excellent scalability and generalizability. The scalability is enabled along three key dimensions: data, model size, and computing power, and generalizability refers to the ability of a model to perform well on new tasks beyond the data it was trained on.

\paragraph{Scalability} 
First, foundation models are trained on massive datasets, frequently comprising hundreds of billions of data samples. The models benefit from this vast data, allowing them to learn very broad patterns. Second, the models themselves are scaled up to huge sizes, frequently with billions or even trillions of parameters. By increasing model capacity, more nuanced representations can be learned from the extensive data. Finally, substantial computing resources are leveraged to train and run these massive models, with thousands of GPUs/TPUs used in parallel computing grids. Together, this facilitates unprecedented scalability, allowing foundation models to achieve capabilities not previously possible with limited data and traditional deep learning models. Their scalability underpins their versatility across a broad array of downstream tasks.

\paragraph{Generalizability} 
Foundation models exhibit strong generalizability due to their training methodology. Specifically, foundation models are pre-trained on diverse datasets in a self-supervised manner, learning broad representations of language, vision, robotics, or other modalities. This differs from traditional supervised learning, where models are tailored to specific labeled datasets. The self-supervised pre-training endows foundation models with extensive world knowledge and versatility. Subsequently, the pretrained models can be adapted and fine-tuned for myriad downstream tasks using relatively little labeled data. Even unseen tasks can be potentially tackled in a zero-shot setting without any downstream data. Their flexibility and wide applicability highlight the generalizability of foundation models across domains. Their generalizability originates from the self-supervised pre-training and enables adaption to new distributions and tasks.

\subsubsection{Dynamic Interactions}
Conventional AI models are restricted to one or several data types as input, and they are even sensitive to the data shape, \emph{i.e.}, 2D or 3D data can often not be processed within the same model due to the limitation of the convolutional operation. Moreover, AI models are always designed for specific tasks, such as image classification or object detection. These limitations constraint the applications of traditional machine and deep learning models, thus, GFMs need to be further developed to enable the handling of multi-modal, multi-type inputs and outputs and so executing flexible tasks.

\paragraph{Multi-Modal Inputs and Outputs}
GFMs aim to represent and understand the physical processes and properties of the Earth’s interior, crust, surface and atmosphere. These models take in multiple types of input data (multimodal inputs) including seismic data, gravity data, heat flow measurements, laboratory rheology data, weather data or land-surface data e.g. from remote sensing. Each data type provides constraints on different aspects of the Earth system. Models can generate multiple types of output (multimodal outputs) including synthetic seismic waves, temperature/pressure fields, surface displacement maps, etc. Handling multi-modal data allows the models to take advantage of complementary constraints from different geophysical observables. This can reduce non-uniqueness and ambiguity compared to using a single data type alone.

\paragraph{Flexible Task}
GFMs aim to represent complex natural processes in Earth system such as mantle convection, crustal deformation, groundwater flow, ocean and atmospheric dynamics, or vegetation. These processes interact in non-linear ways across multiple spatial and temporal scales. GFMs can be taught to flexibly specify new tasks without requiring models to be retrained. For instance, a weather prediction model can be realized by training on historical climate data, and this model can also be helpful for the forecast of rainfall-induced shallow landslides.

\paragraph{Causal Inference}
The integration of GFMs into the exploration of Earth's processes and phenomena heralds a transformative shift in our ability to discern and understand the intricate causal relationships inherent in geoscience knowledge \cite{runge2023causal}. By leveraging data from diverse sources and modalities—ranging from satellite imagery and sensor data to historical climate records and geological samples—GFMs offer a novel approach to modeling the Earth system that is both holistic and nuanced. This comprehensive data integration enables GFMs to capture the complexity and interconnectedness of the Earth system's components. For instance, by analyzing patterns across atmospheric data, oceanic temperatures, and land use changes, GFMs can help researchers identify underlying causal mechanisms driving climate change, such as the feedback loops between greenhouse gas emissions and global warming \cite{massmann2021causal,delforge2022detecting}. Similarly, in the context of natural disasters, GFMs can elucidate the cascading effects of a single event, such as how an earthquake can trigger tsunamis, which in turn affect coastal erosion and sediment transport. The potential of GFMs to elucidate causal relationships in geoscience is grounded in their ability to process and analyze vast datasets that would be unmanageable for traditional analytical methods. Through advanced machine learning techniques, GFMs can detect patterns and correlations within this data, proposing hypotheses about causal links that can be further tested through empirical research or simulation models. This not only accelerates the discovery of new knowledge but also enhances our ability to predict future Earth system changes and inform policy and decision-making. Moreover, the flexibility of GFMs to incorporate data from new sources and update their models in real-time allows for the continuous refinement of our understanding of causal relationships as new evidence emerges. This adaptability is crucial in geoscience, where ongoing changes in the Earth system and technological advancements in data collection methods constantly evolve the landscape of available information. 

\subsection{Geoscience Domain Knowledge}
\label{sub:geoscience_domain_knowledge}

As shown in Fig. \ref{fig:overview}, geoscience encompasses diverse domains spanning the physical makeup and processes of the Earth. Core geoscience domains include geology, geography, geophysics, hydrology, oceanography, meteorology, and planetary science. Each domain has accumulated extensive observations, data, and models of a wide range of Earth system components. For instance, geology involves studying the composition, structure, and history of the Earth's crust and interior, accumulating knowledge on rock types, plate tectonics, mineral resources, and geologic time \cite{warr2022earth}. Meanwhile, meteorology focuses on the atmosphere and atmospheric phenomena such as weather patterns, clouds, and precipitation. Key meteorological knowledge includes air masses, fronts, forecasting models, and climate dynamics \cite{battaglia2020spaceborne}. Moreover, geography tends to explore the spatial patterns and processes of natural phenomena, such as climate change, natural disasters, and biodiversity, as well as human aspects such as urbanization, cultural landscapes, and economic development. Geographers require a diverse set of knowledge, including understanding physical geography, human geography, and technical skills in geographic information systems (GIS), remote sensing, and spatial analysis \cite{mai2023opportunities}. 

Beyond domain-specific knowledge, geosciences also rely on integrative theories and principles regarding Earth systems interactions, such as the carbon cycle, the hydrologic cycle, or interactions between highly nonlinear components of the Earth system that may exhibit abrupt regime transitions  \cite{boers2022theoretical}. Together, robust domain knowledge across the geosciences enables understanding the complex, interconnected processes that shape the Earth and its environments. Accumulating, codifying, and connecting this knowledge is an ongoing endeavor as geoscience domains continue to expand their theoretical and empirical foundations through observation, experimentation, and modeling.

\begin{figure*}[!t]
\centering
\includegraphics[width=0.95\textwidth]{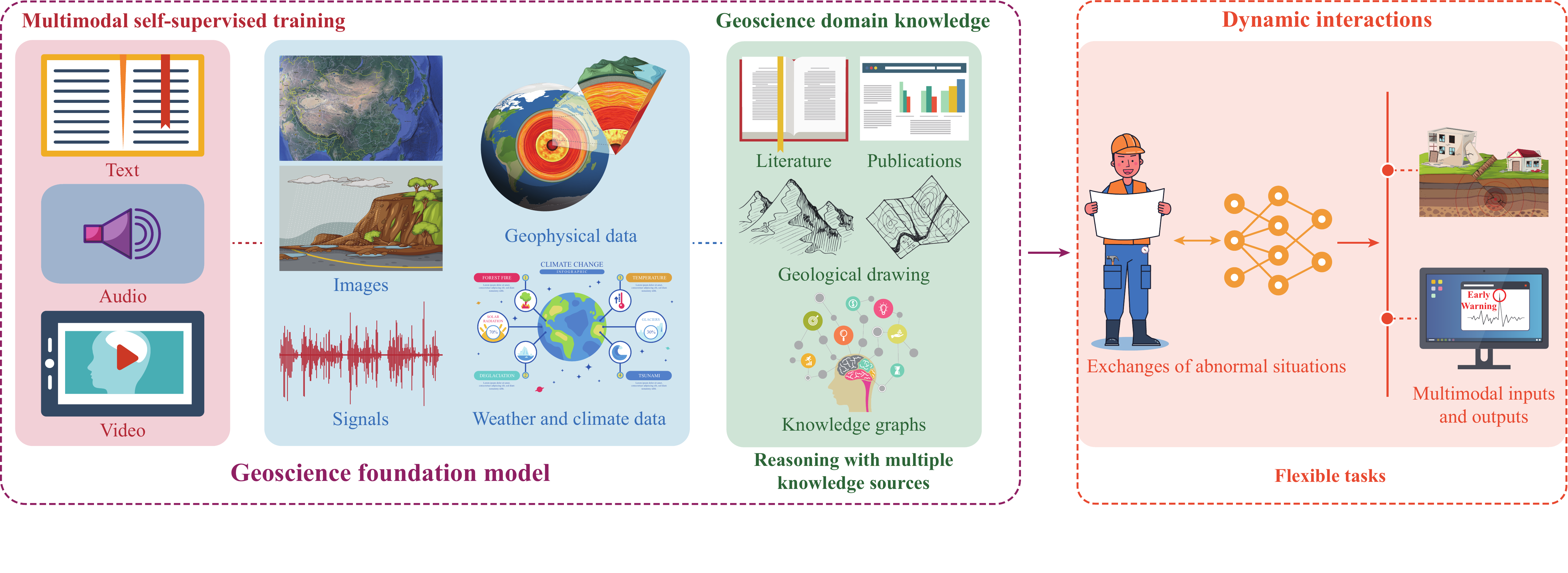}
\caption{\textbf{Overview of the Geoscience foundation model (GFM)}. By leveraging techniques such as self-supervised learning, the GFMs undergo training using multimodal and multidimensional geoscience data models. For enhanced versatility, the data schema can be coupled with language inputs, such as text, voice, or video data. Furthermore, the GFMs require access to diverse sources of geoscientific knowledge to facilitate geoscientific inference tasks, unlocking a plethora of functionalities applicable to downstream applications. Once trained, the GFMs can dynamically perform user-specified tasks in real time. To accomplish this, the model can retrieve contextual information from geological mapping manuscripts, knowledge graphs, databases, and other relevant sources, capitalizing on formal geoscience knowledge to deduce solutions for previously unseen tasks.}
\label{fig:overview}
\end{figure*}

\section{Key Applications of Geoscience Foundation Models}
\label{sec:key_applications}

\begin{figure*}
\centering
\includegraphics[width=0.9\textwidth]{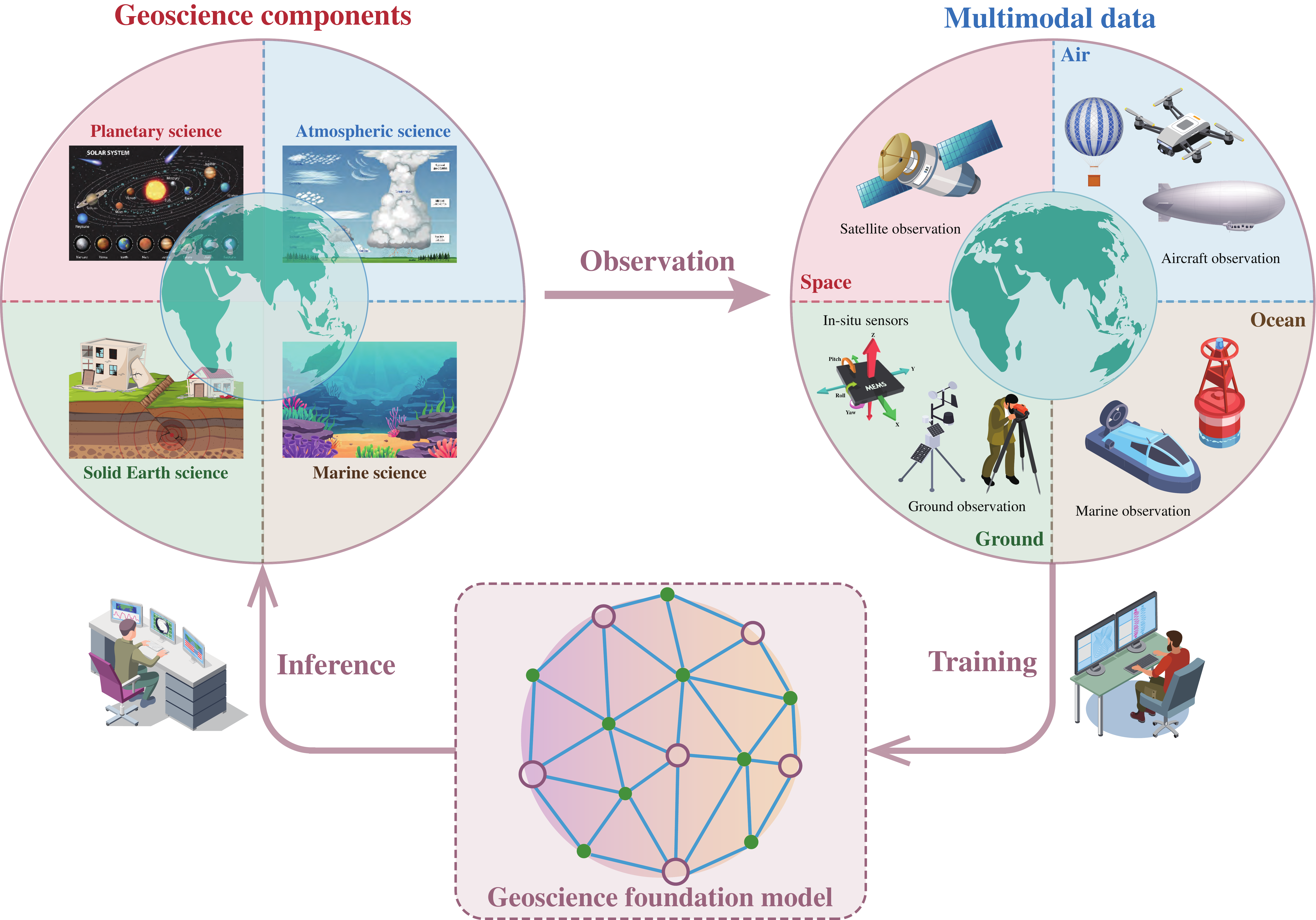}
\caption{\textbf{Illustration of application domain of Geoscience foundation model (GFM)}. On the left side, Earth science is depicted as encompassing four prominent disciplines: solid Earth science, atmospheric science, planetary science, and marine science. On the right side, various observation methods currently utilized in Earth science are represented, including space-based, air-based, ground-based, and ocean-based data collection. These observation methods provide multi-modal and multi-dimensional data that contribute to the field of Earth science. In the middle lies the GFM, which has the capability to integrate, correlate, and analyze the diverse geoscience data obtained from different observation means. By extracting higher-level features and representations from the data, GFMs enable enhanced understanding and insights. Once trained, this model can dynamically perform user-specified tasks in real-time, providing valuable support for Earth science research and applications. }
\label{fig:application_domain}
\end{figure*}

The geoscience field encompasses a diverse range of data types, each fulfilling a distinct role in research and analysis. These data types are pivotal in unraveling the mysteries of our planet and informing various scientific inquiries. As illustrated in Fig. \ref{fig:application_domain}, the diverse array of observational modalities (space-air-ground-ocean) employed in geoscience research each provide unique affordances for characterizing the Earth system. Space-based data collection provides a global perspective, offering comprehensive coverage of the Earth’s surface including vegetation and ecosystems, the atmosphere and oceans. Air-based data collection serves as a bridge between space and terrestrial observations, enabling localized, high-resolution data acquisition. Ground-based data collection involves direct, in-situ measurements, enhancing our comprehension of geographical or geological occurrences. Additionally, ocean-based data collection plays a crucial role in evaluating marine and climatic conditions.

Moreover, GFMs offer significant potential for causal discovery form Apart from multimodal data, considering the diverse observational modalities such as space, air, ground, and ocean. Numerous research questions in Earth sciences are inherently causal, requiring robust analyses to establish the causal relationships among variables from multimodal data \cite{runge2023causal}. By integrating data from various sources and modalities, GFMs enable researchers to uncover causal links between different variables and phenomena, allowing for a more comprehensive and holistic understanding of Earth processes, as it takes into account the interconnectedness of different components of the Earth system. By unraveling the causal mechanisms underlying observed patterns and changes, GFMs contribute to advancing our knowledge of the Earth system and its responses to various drivers and perturbations.

\subsection{Space}
\label{ssub:space}
AI models have demonstrated their capability to integrate the characteristics of the biosphere and ecosystems on Earth, thereby providing valuable assistance in the search and assessment of signs of life on other planets \cite{holzinger2023ai}. We anticipate the development of GFMs that can integrate multimodal and high-dimensional data from the field of planetary science, which aims to obtain more comprehensive information and in-depth insights. GFMs promise to serve as powerful tools in planetary science by facilitating the fusion, correlation, and analysis of data from various modalities. For example, by simultaneously analyzing planetary surface image data, remote sensing data, and radar data, GFMs can facilitate the identification of dependencies and connections among different data sources, thereby enabling more accurate extraction of geomorphic features, identification of geological structures, and prediction of crustal activities. Moreover, with its standardized and adaptable framework for data integration, analysis, and visualization, GFMs are poised to play a pivotal role in unveiling fresh insights and discoveries concerning celestial bodies across our solar system. As space missions become more ambitious, GFMs’ ability to facilitate efficient and collaborative data processing will become increasingly important in advancing our comprehension of planetary processes and environments.

\subsection{Air}
\label{ssub:air}
AI models have gained widespread recognition and utilization in the field of weather forecasting  \cite{bi2023accurate,kurth2023fourcastnet,lam2022graphcast,chen2023fengwu}, with promising advances also regarding extreme weather events such as extreme rainfall \cite{hess2022physically}, long-term droughts, heavy rainfall, typhoons, and strong winds \cite{bi2023accurate}. To enhance the accuracy and timeliness of weather forecasts, there is a need for advanced GFMs that can effectively integrate large volumes of multi-modal meteorological observation data and utilize sophisticated modeling techniques. Such a model would enable meteorologists and forecasters to make more accurate predictions and stakeholders to make more informed evidence-based decisions, leading to improved preparedness, and enhanced understanding of meteorological phenomena. For instance, weather forecasting relies on data from multiple sources, such as satellites \cite{panigrahi2023have}, radars, weather stations, and numerical models. GFMs can serve as a unified platform to integrate and fuse these diverse data streams, providing a comprehensive view of atmospheric conditions. Moreover, the recent success of generative machine learning models such as generative adversarial networks, to bias correct and downscale climate models fields \cite{hess2022physically} suggests that GFM can also play a crucial role for paving the way for improved assessments of climate change impacts. By combining data from different sources, forecasters can make more informed and accurate predictions. GFMs will play a crucial role by providing a standardized semantic representation for weather data, ensuring consistency in data formats, units, and structures. This standardized approach facilitates data sharing, collaboration, and interoperability between different meteorological agencies and research institutions. Moreover, weather forecasting is increasingly reliant on comprehending the intricate interactions between the atmosphere, ocean, land, and cryosphere. GFMs can integrate Earth system models, enabling forecasters to consider complex feedback mechanisms and enhance the accuracy of long-range forecasts. Consequently, GFMs can significantly improve the accuracy and speed of weather forecasting in the future by leveraging their ability to process vast amounts of multimodal data, enable advanced data analytics, and facilitate interdisciplinary collaborations. 

Moreover, the integration of foundation models within the field of remote sensing image segmentation represents a paradigm shift, significantly elevating the precision and efficiency of categorizing complex imagery into distinct, meaningful classes such as vegetation cover, water bodies, and urban infrastructure \cite{hong2023spectralgpt,yan2023ringmo}. Leveraging sophisticated deep learning architectures, these foundation models adeptly navigate the intrinsic variability and intricacy of spatial patterns captured in satellite and aerial imagery. This proficiency is paramount for a spectrum of critical applications, encompassing environmental monitoring (\emph{e.g.}, tracking deforestation, urban expansion), disaster response mechanisms, and agricultural optimization, among others. Foundation models distinguish themselves by their capacity to directly learn high-dimensional feature representations from extensive datasets of annotated images, thereby obviating the need for labor-intensive manual feature extraction. This capability not only streamlines the segmentation workflow but also significantly enhances the adaptability and scalability of remote sensing analysis. Consequently, the adoption of foundation models is not only merely enhancing the operational efficacy of image segmentation tasks, but also concurrently expanding the horizons for innovative research and applications in remote sensing, fostering a more nuanced comprehension of terrestrial phenomena and the intricate interplay between natural and anthropogenic landscapes.

\subsection{Ground}
\label{ssub:ground}
AI models have made significant advancements in the field of geophysics, contributing to various areas such as earthquake prediction and early warning systems, geophysical exploration, and geological risk assessment \cite{bergen2019machine}. Seismic disasters often have significant impacts on society when they occur, yet seismologists currently lack the ability to accurately predict the exact timing of earthquakes. We anticipate the development of comprehensive GFMs that can greatly enhance our understanding of earthquake processes, improve the accuracy of earthquake predictions, and provide valuable information and decision support for earthquake disaster management and risk assessment. For example, earthquake prediction relies on various types of multidimensional seismic data, such as seismic waveforms, seismic records, and underground geological data. Advanced GFMs excel at processing high-dimensional data and integrating multiple modalities of seismic information, thereby enabling more comprehensive and accurate earthquake prediction outcomes. Furthermore, it is essential to acknowledge that although GFMs can offer substantial contributions to earthquake prediction, this field remains exceptionally challenging due to the intricate nature of seismic processes. Continuous research endeavors, advancements in data collection methodologies, and collaborative initiatives are pivotal in further enhancing the accuracy and efficiency of earthquake prediction. Nevertheless, GFMs hold tremendous potential in revolutionizing earthquake forecasting and mitigating the potential impact of seismic events on communities and critical infrastructure.

\subsection{Ocean}
\label{ssub:ocean}
AI models enable holistic monitoring, modeling, and forecasting of conditions in the near-surface ocean through multi-sensor data analysis, digital twin simulations, and long-term observations at multi-scale \cite{sonnewald2021bridging}. However, compared with the near-surface ocean, the intermediate ocean at a depth from 100 to 1000 m is confronted with a series of challenges primarily stemming from the scarceness of observational data, which has emerged as a significant impediment constraining the progressive evolution of marine science and data-driven modelling in oceanography. We anticipate the development of a GFM featuring real-time, immersive, and highly sensitive dynamic interactive visualization capabilities for three-dimensional oceanic data. This model holds the promise of gradually unraveling the enigma surrounding the intermediate and deep oceans, thereby ushering future marine science into a transformative new phase of exploration and understanding. For instance, advanced GFMs have the potential to catalyze a transformation in mid-ocean remote sensing technology, offering the capability for autonomous, long-term monitoring of poorly explored mid-ocean regions. This can be achieved through intelligent fusion and analysis of diverse subsurface, environmental, and biological data streams from distributed, next-generation ocean observation platforms, thereby revolutionizing the way mid-ocean regions are studied and understood. Moreover, GFMs play a transformative role within the framework of Deep Blue AI \cite{ma2021visual}, facilitating advanced data analysis, predictive modeling, and well-informed decision-making to elevate the quality of marine science research, exploration, and conservation endeavors, thereby helping to unraveling the enigma of one of Earth’s least-explored yet profoundly significant frontiers. 

\section{Key Techniques of Constructing Geoscience Foundation Models}
\label{sec:key_techniques}
\begin{figure*}
\centering
\includegraphics[width=0.9\textwidth]{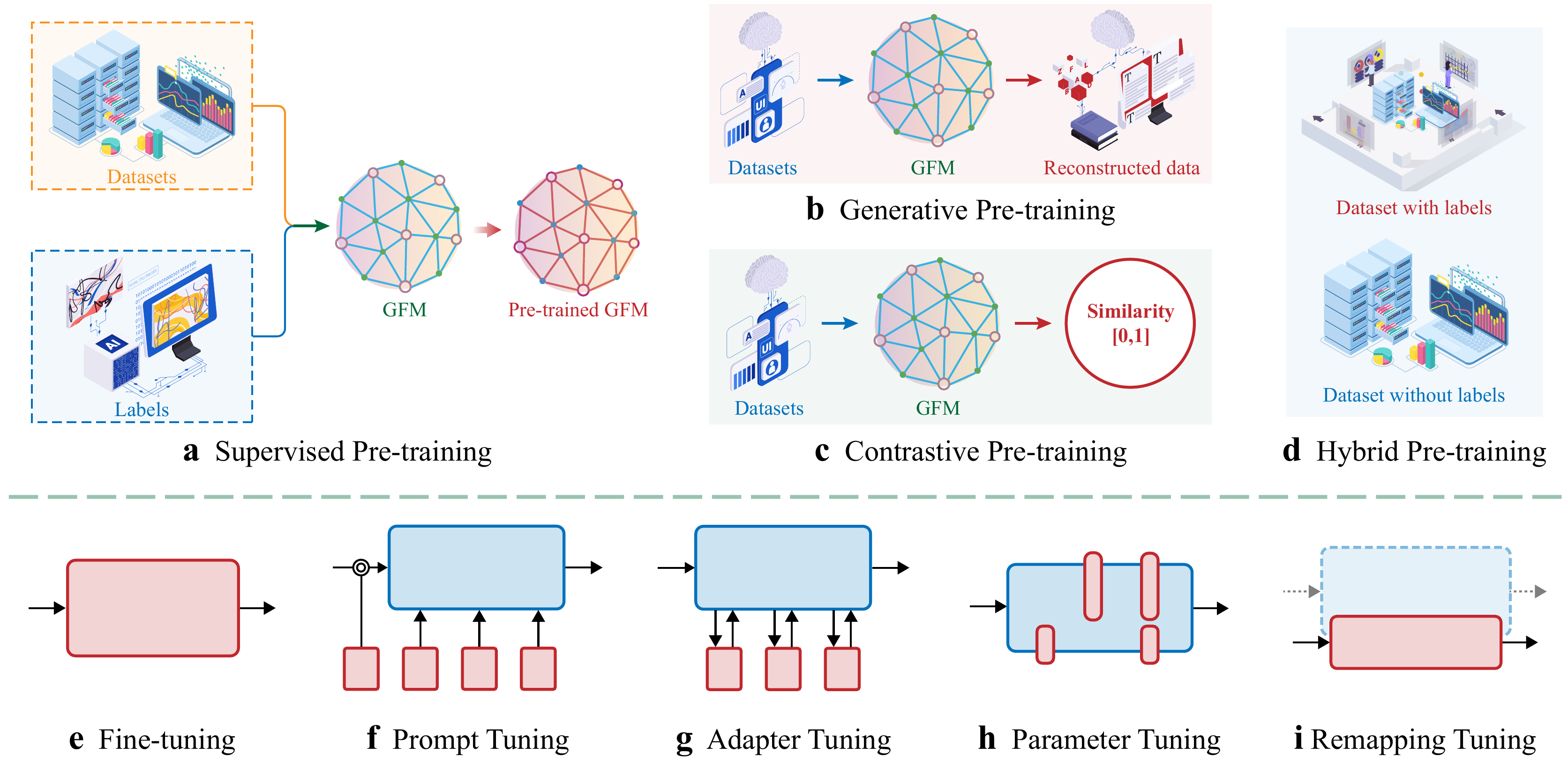}
\caption{\textbf{Key Techniques of Constructing Geoscience Foundation Models, including Pre-training (a-d) and Adaptation (e-i).} \textbf{a.} Supervised Pre-training, training with large labeled data. \textbf{b.} Generative Pre-training, training by reconstructing data. \textbf{c.} Contrastive Pre-training, training by contrastive learning using similarity. \textbf{d.} Hybrid Pre-training, training with labeled and unlabeled data. \textbf{e.} Fine-tuning, fine-tuning GFMs using domain specific datasets. \textbf{f.} Prompt Tuning, fine-tuning GFMs using prompts. \textbf{g.} Adapter Tuning, fine-tuning GFMs using additional trainable parameters. \textbf{h.} Parameter Tuning, fine-tuning GFMs by modifying specific parameters. \textbf{i.} Remapping Tuning, fine-tuning GFMs by knowledge transfer.}
\label{fig4}
\end{figure*}

In this section, we discuss the key techniques for constructing geoscience foundation models (GFMs), including transformers, pre-training and adaptation. 
As shown in Fig. \ref{fig4}, pre-training and adaptation are two critical stages in the development and application of foundation models, where the former involves training a model on a large, diverse dataset to learn general-purpose representations, and the latter involves fine-tuning the pre-trained model on a specific task or dataset. The pre-training stage is particularly important for foundation models, as it enables the model to learn general-purpose representations that can be adapted to a wide range of tasks. The adaptation stage is equally important, as it allows the model to be fine-tuned for specific tasks or datasets, thereby improving its performance on those tasks. 

\subsection{Transformers}
The Transformer architecture, introduced by Vaswani \et \cite{vaswani2017attention}, represents a significant leap forward in neural network design for natural language processing (NLP). At its core, the Transformer's power stems from its self-attention mechanisms, which enable effective modeling of long-range dependencies and contextual relationships in language \cite{yang2022transformers}. This approach addresses limitations observed in traditional sequential models such as recurrent neural networks (RNNs) and convolutional neural networks (CNNs), which often struggle with processing long sequences \cite{yu2019review, zhang2016relation}. By calculating correlations between each token and all others in the input sequence, the Transformer mitigates challenges faced by conventional models, making it a versatile tool for various NLP applications. For more details, please refer to the related literatures \cite{lin2022survey,han2022survey,xu2023multimodal}.

\subsubsection{Vanilla Transformer}
\begin{figure}
\centering
\includegraphics[width=0.45\textwidth]{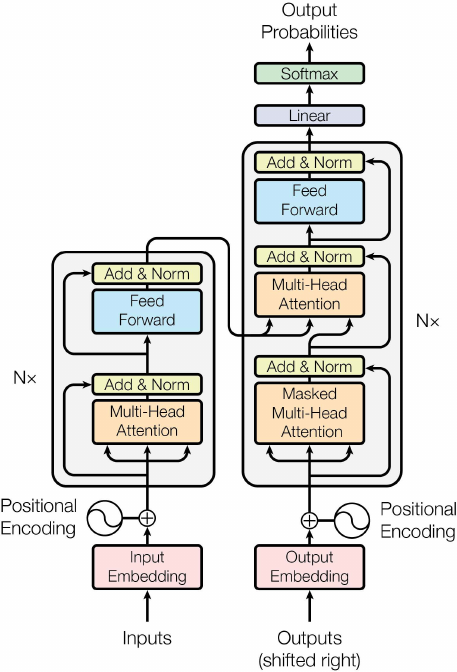}
\caption{The Transformer model architecture \cite{vaswani2017attention}.}
\label{fig:transformer}
\end{figure}

The Transformer's architecture is primarily composed of an encoder and a decoder, each featuring multi-head self-attention mechanisms and position-wise feed-forward networks \cite{vaswani2017attention}. In the encoder, input sequences are projected into multiple vector spaces, undergoing a series of linear transformations, attention computations, and concatenation operations to generate context-aware output vectors. The decoder uses these context vectors to generate the output sequence, with an additional attention layer to focus on relevant parts of the input, similar to the attention mechanism in Seq2Seq models \cite{sutskever2014sequence}. This structure, as illustrated in Fig. \ref{fig:transformer}, allows the Transformer to process extended text sequences with equal consideration given to all positions, offering advantages over recurrent architectures in terms of parallelization and long-range dependency modeling \cite{wolf2020transformers, kalyan2021ammus, yang2023transformer,khan2022transformers}. 

\subsubsection{Visual Transformer}
The Transformer architecture has revolutionized NLP with remarkable success. Its impact has since expanded into computer vision, demonstrating impressive capabilities across various tasks such as image recognition, classification, and segmentation. The Transformer has emerged as a versatile and scalable framework, offering significant advantages in training efficiency compared to traditional methods. Its flexibility allows for implementation as a standalone architecture or in combination with CNNs to achieve superior results in visual tasks. Typical examples include the vision transformer (ViT) \cite{dosovitskiy2020image}, which has garnered widespread attention for its convolution-free design and exceptional ability to model long-range dependencies. 

\begin{figure}
\centering
\includegraphics[width=0.45\textwidth]{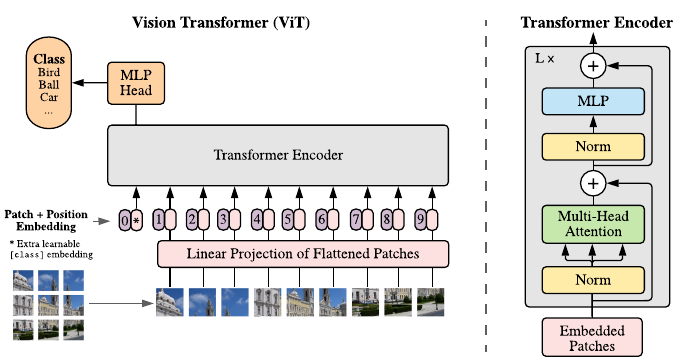}
\caption{The framework of vision transformer (ViT) \cite{dosovitskiy2020image}.}
\label{fig:vit}
\end{figure}

ViT architecture \cite{dosovitskiy2020image} is illustrated in Fig. \ref{fig:vit}. Initially, the input image is segmented into $16 \times 16$ patches, which are then flattened and linearly projected. These processed patches, combined with positional encodings, are fed into the Transformer encoder. The encoded features are subsequently passed through an MLP head for classification. The transformer encoder primarily utilizes the positional encoding structure proposed in the original Transformer model. ViT represents a pioneering application of the Transformer architecture to computer vision tasks \cite{khan2022transformers}. Its convolution-free design and exceptional ability to model long-range dependencies have garnered significant attention from researchers. Building upon ViT's foundation, several variants have emerged, including DeiT \cite{touvron2021training}, TNT \cite{han2021transformer}, and PVT \cite{wang2021pyramid}. These iterations have further enhanced performance by incorporating advanced techniques such as knowledge distillation, hierarchical token mixing, and pyramid-style vision transformers.

\subsubsection{Vision-Language Transformer}
Vision-language transformers bridge the gap between computer vision and natural language processing. These models leverage the powerful Transformer architecture to simultaneously process and understand both visual and textual information, enabling a wide range of multimodal tasks. By learning joint representations of images and text, vision-language transformers can perform complex operations such as visual question answering, image captioning, and cross-modal retrieval with remarkable efficiency and accuracy. Most representative models include VisualBERT \cite{li2019visualbert}, Uniter \cite{chen2020uniter}, and OSCAR \cite{li2020oscar}. These models integrate BERT for text encoding and Faster R-CNN for object detection, enabling the joint processing of visual and textual inputs through multiple transformer layers. 

\begin{figure}
\centering
\includegraphics[width=0.45\textwidth]{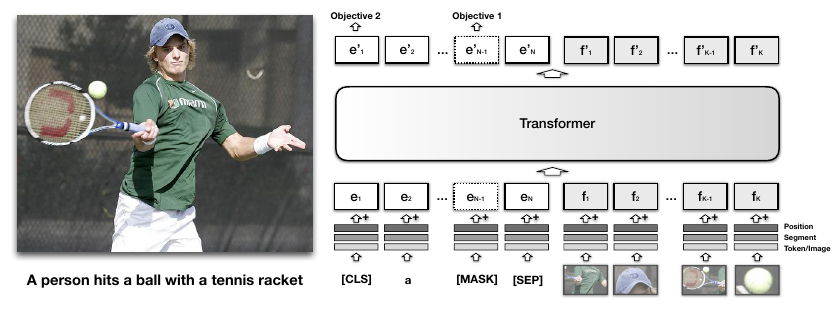}
\caption{The architecture of VisualBERT\cite{li2019visualbert}.}
\label{fig:visualbert}
\end{figure}

VisualBERT \cite{li2019visualbert} pioneered the implicit alignment of textual elements with image regions using self-attention mechanisms. This model integrated BERT \cite{devlin2018bert} for NLP with a pretrained Faster-RCNN \cite{ren2015faster} for object detection. The approach treated the original text and image features extracted from object proposals as unordered input tokens, processing them jointly through multiple Transformer layers to capture complex cross-modal associations (as shown in Fig. \ref{fig:visualbert}). Following VisualBERT's innovative approach, several VLMs emerged, including Uniter \cite{chen2020uniter}, OSCAR \cite{li2020oscar}, and InterBert \cite{lin2020interbert}. These models similarly employed BERT as their text encoder and Faster-RCNN for object proposal generation, further advancing the modeling of vision-language interactions.

\subsection{Pre-training} 

Pre-training involves training a foundation model on a large, diverse dataset. The primary goal is to learn a broad range of patterns, knowledge, and skills that can be universally applied across various tasks. Pre-training methods for GFM include supervised pre-training, self-supervised pre-training, and hybrid pre-training, where self-supervised pre-training involves generative and contrastive pre-training \cite{liu2021self}. 

\subsubsection{Supervised pre-training} 

Supervised learning has been a cornerstone in the development of deep learning models, driving success in both vision and language tasks. In computer vision, models including AlexNet \cite{krizhevsky2012imagenet}, VGG \cite{simonyan2014very}, and ResNet \cite{he2016deep} achieved breakthrough performance through supervised pre-training on large-scale datasets such as ImageNet \cite{deng2009imagenet}, JFT-300M \cite{sun2017revisiting}, and Kinetics \cite{kay2017kinetics}. Similarly, in natural language processing, early transformer-based models such as BERT \cite{devlin2018bert}, RoBERTa \cite{liu2019roberta}, and ALBERT \cite{lan2019albert} utilized supervised pre-training on tasks such as masked language modeling and sentence ordering prediction. As the field progressed, researchers explored larger-scale and more diverse pre-training approaches, such as Gato's \cite{reed2022generalist} multi-modal, multi-task learning, and multi-label learning in vision models scaling up to 22 billion parameters \cite{dehghani2023scaling}. In the biomedical domain, BioBERT \cite{lee2020biobert} specializes in medical text analysis, while ELECTRA \cite{clark2020electra} employs a novel discriminator-based approach for efficient language model pre-training. VideoBERT \cite{sun2019videobert} extends to multimodal learning for video understanding. This evolution in pre-training strategies highlights the ongoing quest for more efficient and generalizable learning paradigms in AI, balancing the need for task-specific knowledge with broader applicability to avoid overfitting.

\subsubsection{Self-supervised pre-training}
Self-supervised learning (SSL) has become a cornerstone in the development of foundation models (FMs), enabling them to acquire rich features and representations from vast amounts of unlabeled data \cite{jing2020self}, \cite{chen2023self}, \cite{shen2023firerisk}. This approach, often referred to as the "dark matter" of intelligence, circumvents the limitations of supervised learning by leveraging a "semiautomatic" process to derive "labels" directly from the data, significantly reducing time and resource costs \cite{chen2023self}. SSL has achieved remarkable success across various domains of deep learning, particularly in natural language processing with models such as BERT \cite{devlin2018bert} and GPT-3 \cite{brown2020language}, and in computer vision with MAE \cite{he2022masked} and DINOv2 \cite{oquab2023dinov2}. The SSL workflow typically involves defining a pretext task based on unlabeled inputs to generate meaningful representations \cite{jing2020self}, \cite{shen2023firerisk}, \cite{friedman2009elements}, \cite{goodfellow2016deep}. These pretext tasks, which can be prediction-based, context-based, or generation-based, form the basis of the pre-training phase where the objective function is learned from unlabeled data \cite{friedman2009elements}. The resulting representations are then transferred as initial weights for fine-tuning on specific downstream tasks. SSL methods can be broadly categorized into three approaches based on their pretext tasks: generative, contrastive, and predictive \cite{jing2020self}, \cite{wang2022self}. This “pre-training, fine-tuning” paradigm has proven highly effective, allowing FMs to leverage the power of unlabeled data before being adapted to specific tasks with labeled data, often matching or surpassing purely supervised approaches in performance.

\begin{figure}
  \centering
  \resizebox{0.48\textwidth}{!}{
  \begin{forest}
  for tree={
    font=\footnotesize,
    align=center,
    child anchor=west,
    parent anchor=east,
    grow'=0,
    edge={-Stealth},
    edge path={
      \noexpand\path [draw, \forestoption{edge}] (!u.parent anchor) .. controls +(0.7,0) and +(-0.7,0) .. (.child anchor)\forestoption{edge label};
    },
    l sep+=0.7cm,
    s sep+=0.2cm,
    tier/.wrap pgfmath arg={tier #1}{level()},
  }
  [Pre-\\training, fill=gray!60, text=white, rounded corners
    [Supervised, fill=blue!80, text=white, rounded corners
      [CNNs
        [AlexNet \cite{krizhevsky2012imagenet}] 
        [VGG \cite{simonyan2014very}]
        [ResNet \cite{he2016deep} ]
      ]
      [Transformers
        [BERT \cite{devlin2018bert}] 
        [RoBERTa \cite{liu2019roberta}]
      ]
    ]
    [SSL, fill=blue!80, text=white, rounded corners
      [Generative
        [GAN
          [BigGAN \cite{brock2018large}]
          [SRGAN \cite{ledig2017photo}]
        ]
        [Autoencoder
          [VAE \cite{kingma2013auto}]
          [MAE \cite{he2022masked}]
        ]
      ]
      [Contrastive
        [Negative sampling
          [SimCLR \cite{chen2020simple}]
          [MoCo \cite{he2020momentum}]
        ]
        [Clustering
          [DeepCluster \cite{caron2018deep}]
          [SwAV \cite{caron2020unsupervised}]
        ]
        [Knowledge\\ distillation
          [BYOL \cite{grill2020bootstrap}]
          [DINO \cite{caron2021emerging}]
        ]
        [Redundancy\\ reduction
          [Barlow Twins \cite{zbontar2021barlow}]
        ]
      ]
      [Predictive
        [Spatial
          [CFN \cite{pathak2016context}]
          [RotNet \cite{gidaris2018unsupervised}]
        ]
        [Spectral
          [Colorful Image\\ Colorization \cite{zhang2016colorful}]
        ]
      ]
    ]
    [Hybrid, fill=blue!80, text=white, rounded corners
      [ELECTRA \cite{clark2020electra}]
      [DALL-E \cite{ramesh2022hierarchical}]
    ]
  ]
  \end{forest}
  }
\caption{Taxonomy of the pre-training methods with examples.}
\label{fig:ssl-taxonomy}
\end{figure}

\paragraph{Generative pre-training}
Generative methods in self-supervised learning focus on learning representations through the process of reconstructing or generating input data. These approaches seek to model the underlying data distribution, thereby capturing intrinsic statistical properties and dependencies within the data \cite{goodfellow2020generative}. By doing so, generative methods can implicitly extract meaningful features and structures without the need for explicit labels. This capability makes them particularly valuable in scenarios where labeled data is scarce or expensive to obtain. Common implementations of generative methods often employ sophisticated models such as autoencoders, which compress and then reconstruct input data, or generative adversarial networks (GANs), which pit a generator against a discriminator in a competitive learning process.

Autoencoders form the foundation of many generative methods, consisting of an encoder network that maps input data to a latent space representation and a decoder network that reconstructs the data from this latent space. This structure has spawned several variants, notably the variational autoencoder (VAE) \cite{kingma2013auto}, which incorporates probabilistic modeling by assuming a prior distribution $P(Z)$ over the latent space and modeling the conditional distribution $P(X|Z)$. In VAEs, the encoder approximates the posterior distribution $P(Z|X)$ by inferring the model $Q(Z|X)$. Recent applications of generative methods include information recovery tasks such as inpainting \cite{pathak2016context}, where a network's context encoder restores missing pixel values based on surrounding context. This concept has evolved into masked autoencoders (MAEs) \cite{he2022masked}, which apply random masks to input image patches and reconstruct the missing pixels. The process of accurately reconstructing each pixel requires the model to develop a comprehensive understanding of the various objects and components in the image, resulting in learned feature representations that prove valuable for a wide range of downstream tasks.

GANs \cite{goodfellow2020generative} represent a powerful approach in generative methods, consisting of two competing networks: a generator $G: Z \rightarrow X$ that synthesizes fake samples from random noise in the latent space, and a discriminator $D: X \rightarrow [0, 1]$ that attempts to distinguish between real and fake samples. This adversarial setup has proven effective for tasks such as image generation and super-resolution. Notable advancements include BigGAN \cite{brock2018large}, which introduces innovations sucha s a large-scale GAN architecture, conditional batch normalization, category conditioning, and orthogonal regularization, enabling the generation of high-quality, diverse, and category-specific images. In the domain of image enhancement, SRGAN \cite{ledig2017photo} has demonstrated remarkable capabilities in super-resolution tasks, learning from a large dataset of downsampled images to recover high-resolution textures and effectively upscale low-resolution inputs. These applications showcase the potential of GANs not just for generating new images, but also for enhancing existing ones, pushing the boundaries of synthetic image generation and opening up numerous possibilities in image processing and computer vision tasks.

\paragraph{Contrastive pre-training}
Contrastive learning methods aim to learn representations by maximizing similarity between related samples and minimizing it between unrelated ones. This approach uses a contrastive objective function to bring similar samples closer in feature space while separating dissimilar ones. The key advantage of contrastive learning is its ability to extract meaningful representations without explicit labels, leveraging the data's inherent structure. These methods typically fall into four categories: negative sampling, clustering, knowledge distillation, and redundancy reduction. 

Negative sampling in contrastive learning involves creating dissimilar examples to anchor samples, aiming to differentiate positive and negative instances in the representation space. This approach helps learn valuable features while avoiding model collapse. Notable implementations include SimCLR \cite{chen2020simple}, which uses a CNN feature extractor with diverse data augmentations, and the MoCo series \cite{he2020momentum}. MoCo employs a momentum-based update strategy and compares samples against a large queue of negative examples. MoCo and its variants (MoCov2 \cite{chen2020improved}, MoCov3 \cite{chen2021empirical}) have shown strong performance in self-supervised learning tasks, demonstrating the effectiveness of negative sampling in developing robust and discriminative representations. These methods have significantly advanced the field of contrastive learning, providing powerful tools for learning from unlabeled data in various computer vision tasks.

The clustering approach in contrastive learning groups similar samples unsupervised. DeepCluster \cite{caron2018deep} exemplifies this method by first clustering images and then training a CNN to recognize these assignments. Another notable example is SwAV \cite{caron2020unsupervised}, which introduces the concept of grouping similar representations. SwAV generates multiple views of an image through data augmentation, clusters these views based on similarity, and encourages similar representations within clusters while differentiating between clusters. These methods demonstrate how unsupervised clustering can effectively create meaningful groupings in the feature space, enabling the model to learn rich representations without explicit labels. By leveraging the inherent structure in the data, clustering approaches in contrastive learning contribute to developing more robust and adaptable models for various computer vision tasks.

Knowledge distillation in self-supervised learning (SSL) transfers knowledge from a high-performing teacher model to a student model. Notable examples include BYOL \cite{grill2020bootstrap} and the DINO series \cite{caron2021emerging}, \cite{oquab2023dinov2}. BYOL predicts representations of augmented views, using an online network as the student and a momentum-updated target network as the teacher. DINO combines transformer architecture with consistency maximization between image views, where one view predicts the teacher network's output for another. DINOv2 enhances this approach by adding a clustering objective, encouraging semantically similar samples to cluster in the learned representation. These methods demonstrate how knowledge distillation can effectively leverage pretrained models to guide the learning of robust, fine-grained representations in SSL, advancing the field of unsupervised visual representation learning.

Redundancy reduction in contrastive learning aims to minimize redundant information in learned representations, focusing on capturing essential and discriminative features. A notable example is the Barlow twins model \cite{zbontar2021barlow}, which optimizes the cross-correlation matrix of representations. This approach maximizes cross-correlation between different views of the input while minimizing diagonal elements, effectively encouraging the model to encode useful, non-redundant information. By balancing the capture of statistical dependencies with redundancy reduction, this method enables the learning of more efficient and informative representations. This approach contributes to the development of more compact and effective feature representations, potentially improving model performance and generalization across various computer vision tasks.

\paragraph{Predictive pre-training}
Predictive methods in self-supervised learning aim to learn useful representations through various prediction tasks, encompassing both spatial and spectral aspects of images. Spatial-based approaches include relative position prediction, where CNNs are trained to predict the positions of randomly sampled image patches \cite{noroozi2016unsupervised}; image puzzles, which involve reconstructing images from shuffled patches to learn object part relationships and shapes; geometric transformation recognition, such as RotNet \cite{gidaris2018unsupervised}, which identifies rotation angles of images; and counting tasks that train models to enumerate visual primitives. Spectral-based methods, exemplified by image colorization (IC) tasks \cite{zhang2016colorful}, focus on automatically adding realistic colors to grayscale images. These diverse approaches encourage models to learn rich, contextual representations by solving pretext tasks that leverage the inherent structure and properties of images. For instance, the Jigsaw puzzle approach in CFN \cite{noroozi2016unsupervised} trains networks to understand spatial arrangements of object parts, while IC tasks help models learn complex color distributions and object relationships. By engaging with these challenging tasks, models develop robust feature representations that prove valuable for downstream applications such as classification, detection, and segmentation, demonstrating the effectiveness of predictive methods in unsupervised visual learning.

\subsubsection{Hybrid pre-training}

Hybrid pre-training approaches have significantly advanced the development of large AI models across multiple domains, combining diverse learning techniques to leverage their strengths and address limitations. In large language models (LLMs), examples include ELECTRA's \cite{clark2020electra} generator-discriminator setup and BART's \cite{lewis2019bart} combination of bidirectional encoding and autoregressive decoding. Large vision models (LVMs) often pair contrastive learning with supervised tasks, as seen in SimCLR \cite{chen2020simple} and MAE \cite{he2022masked}. Large vision-language models (LVLMs) such as CLIP \cite{radford2021learning} and DALL-E \cite{ramesh2022hierarchical} utilize cross-modal contrastive learning, combining vision and language inputs. These approaches typically involve self-supervised pre-training on large unlabeled datasets followed by supervised fine-tuning. By integrating multiple pre-training objectives and architectures, hybrid approaches enable models to capture diverse aspects of their domains – from language understanding and generation to visual feature extraction and cross-modal reasoning – resulting in improved performance across a wide range of downstream tasks.

\subsection{Adaptation} 
Pre-trained GFMs can be adapted to specific tasks or datasets through fine-tuning, prompt tuning, adapter tuning, parameter tuning, and remapping tuning. For more tuning strategies, interested readers can refer to the literatures \cite{lialin2023scaling,yu2024visual,xing2024survey,zheng2023learn}.

\begin{figure}
  \centering
  \resizebox{0.48\textwidth}{!}{
  \begin{forest}
  for tree={
    font=\footnotesize,
    align=center,
    child anchor=west,
    parent anchor=east,
    grow'=0,
    edge={-Stealth},
    edge path={
      \noexpand\path [draw, \forestoption{edge}] (!u.parent anchor) .. controls +(0.7,0) and +(-0.7,0) .. (.child anchor)\forestoption{edge label};
    },
    l sep+=0.7cm,
    s sep+=0.2cm,
    tier/.wrap pgfmath arg={tier #1}{level()},
  }
  [Adaptation, fill=gray!60, text=white, rounded corners
    [Fine \\ tuning, fill=blue!80, text=white, rounded corners
    ]
    [Prompt\\tuning, fill=blue!80, text=white, rounded corners
      [Vision-driven
        [VPT \cite{jia2022visual}
        DePT \cite{gao2022visual}
        PViT \cite{herzig2024promptonomyvit}]
      ]
      [Language-driven
        [CoOp \cite{zhou2022learning}
        PLOT \cite{chen2022plot}]
      ]
      [Vision-language
        [UPT \cite{zang2022unified}
        DPT \cite{xing2023dual}
        TPT \cite{shu2022test}]
      ]
    ]
    [Adapter \\Tuning, fill=blue!80, text=white, rounded corners
      [Sequential Adapter
        [Res-adapt \cite{rebuffi2017learning}
        DAN \cite{rosenfeld2018incremental}
        LST \cite{sung2022lst}
        Conv-\\Adapter \cite{chen2024conv}]
      ]
      [Parallel Adapter
        [ViT-Adapter \cite{dosovitskiy2020image}
        PESF-KD \cite{rao2023parameter}
        AdaptMLP \cite{chen2024conv}]
      ]
      [Mix Adapter
        [PATT \cite{yu2022towards}
        ETT \cite{xu2023exploring}
        PALT \cite{wu2022pruning} 
        VQT \cite{tu2023visual}]
      ]
    ]
    [Parameter\\ Tuning, fill=blue!80, text=white, rounded corners
      [Weight Part
        [LoRA \cite{hu2021lora} DyLoRA \cite{valipour2022dylora}]
      ]
      [Bias Part
        [Bitfit \cite{zaken2021bitfit}
        AdapterBias \cite{fu2022adapterbias}]
      ]
      [Weight and Bias Parts
        [SSF \cite{lian2022scaling}]
      ]
    ]
    [Remapping\\ Tuning, fill=blue!80, text=white, rounded corners
      [Knowledge Distillation
        [\cite{buciluǎ2006model,hinton2015distilling}]
      ]
      [Weight Remapping
        [Net2Net \cite{chen2015net2net} EAS \cite{cai2018efficient}]
      ]
      [Architecture Remapping
        [DARTS \cite{liu2018darts}]
      ]
    ]
  ]
  \end{forest}
}
\caption{Taxonomy of the adaptation methods with examples.}
\label{fig:adaptation}
\end{figure}

\subsubsection{Fine-Tuning}
Model fine-tuning involves updating the weights of a pre-trained model to adapt it to a specific task or dataset. This process typically occurs after the initial pre-training phase and aims to leverage the general knowledge captured by the pre-trained model while specializing it for a particular application. The weights of the model are updated to minimize the prediction error on the specific task or dataset, often using a smaller learning rate than in pre-training to preserve some of the valuable representations learned during pre-training. Model fine-tuning is particularly useful when there is a small, labeled dataset available for adaptation, as it allows the model to benefit from the knowledge acquired from a large dataset during pre-training while adjusting to the nuances of the target task.

\subsubsection{Prompt Tuning}
\begin{figure}
\centering
\includegraphics[width=0.45\textwidth]{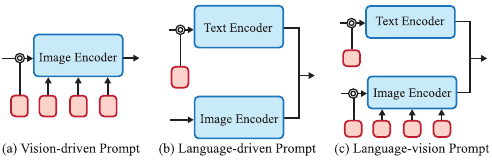}
\caption{Three different types of prompt methods \cite{yu2024visual}.}
\label{fig:prompt}
\end{figure}

Prompt-based tuning initially introduced in NLP, offers an efficient method for adapting foundation models to downstream language tasks. This approach diverges from the traditional “pre-training, fine-tuning” paradigm, which initializes and optimizes the weight parameters of pre-trained models using task-specific loss functions. Instead, prompt-based learning employs textual prompts to reformulate various downstream tasks within the framework of the original pre-trained task. Drawing inspiration from these NLP techniques, prompt tuning has subsequently been applied to the field of computer vision. In this context, vision prompt tuning can be categorized into three distinct approaches, including vision-driven prompts, language-driven prompts, and vision-language prompts, as shown in Fig. \ref{fig:prompt}. 
\begin{itemize}
  \item \emph{Vision-driven Prompts}: Vision-driven prompt tuning is a widely used, efficient method for adapting pre-trained vision models to new tasks while maintaining their strong generalization capabilities. 
  This approach comes in two main flavors: directly modifying input images or using special sub-networks to create vision prompts. In the first method, researchers add learnable elements to input images, tweaking them to better match what the model was originally trained on. The second method involves creating small, specialized networks that generate prompts to guide the main model in handling new tasks. Both strategies aim to make the most of a pre-trained model's existing knowledge, allowing it to tackle new challenges with minimal adjustments. This technique has gained significant attention in recent research due to its effectiveness in preserving a model's general capabilities while adapting it for specific purposes. VPT \cite{jia2022visual}, DePT \cite{gao2022visual}, and PViT \cite{herzig2024promptonomyvit} are some of the most prominent examples of vision-driven prompt tuning.
  \item \emph{Language-driven Prompts}: Recently, large-scale vision-language models are pre-trained by extensive image-text pairs and focus on open-world visual concepts. Following this ideology of prompt learning in NLP, most existing works tend to transfer large-scale vision-language models into various downstream vision tasks via designing appropriate language-driven prompts. CoOp \cite{zhou2022learning},  PLOT \cite{chen2022plot} are two examples.
  \item \emph{Vision-language Prompts}: Vision-driven and language-driven prompts have been explored to simultaneously modify the vision and text inputs for pre-trained vision-language models, thus transferring the discrimination and generalization ability of pre-trained vision-language models thanks to effectively aligning visual and textual embeddings. UPT \cite{zang2022unified}, DPT \cite{xing2023dual}, TPT \cite{shu2022test} can be referred.
\end{itemize}

\subsubsection{Adapter Tuning}
Adapter-based tuning represents a class of techniques that introduce additional trainable parameters into a frozen pre-trained model, facilitating efficient learning for downstream tasks. These methods were initially introduced in the field of NLP by Houlsby \et \cite{houlsby2019parameter} as a means of achieving parameter-efficient transfer learning (PETL). While efficient adaptation has been extensively explored in NLP, it has received comparatively less attention in the domain of computer vision. Early efforts to develop adaptive methods for computer vision focused on incremental learning \cite{rosenfeld2018incremental} and domain adaptation \cite{rebuffi2017learning,rebuffi2018efficient}. More recently, adapters have gained traction across various domains and have been successfully applied to computer vision tasks. These adapters offer a lightweight alternative to extensive model fine-tuning, providing a more efficient approach to task-specific adaptation while preserving the pre-trained model's general knowledge.
\begin{itemize}
  \item \emph{Sequential Adapter}: Sequential adapter is a technique that enhances a sequential forward network by inserting additional parameters. It involves a series of operations: a linear down-projection, followed by a non-linear activation function, then an up-projection, and finally a residual connection. This approach is typically implemented following the multi-head attention layer, the feed-forward layer, or both, with the aim of improving the model's overall effectiveness. Res-adapt \cite{rebuffi2017learning}, DAN \cite{rosenfeld2018incremental}, LST \cite{sung2022lst}, and Conv-Adapter \cite{chen2024conv} adopt sequential adapter strategies. 
  \item \emph{Parallel Adapter}: Parallel adapter is an alternative to the traditional sequential adapter. In this variant, activations flow through the module layer simultaneously with the adapted sub-layer (such as the feed-forward or attention layer), rather than following the conventional sequential order of operations. ViT-Adapter \cite{chen2022vision}, PESF-KD \cite{rao2023parameter}, AdaptMLP \cite{chen2022adaptformer} are examples of parallel adapter techniques. 
  \item \emph{Mix Adapter}: Mix adapter introduces new parameters in different positions with mixed architecture, \emph{i.e.}, the multi-head attention blocks in each transformer layer. PATT \cite{yu2022towards}, ETT \cite{xu2023exploring}, PALT \cite{wu2022pruning}, VQT \cite{tu2023visual} explored the usage of mix adapter strategies.
\end{itemize}

\subsubsection{Parameter Tuning}
Parameter-based tuning is a methodology that involves direct and substantial modifications to the parameters of a pre-trained model, specifically altering its weights or biases. In a given layer, the weight-term can be multiplied by the feature map, while the bias-term can be added to it. This approach can be categorized based on which parameter components are adjusted: the weight component, the bias component, or both simultaneously. The techniques employed in parameter-based tuning can be further classified into two primary categories: weight part, bias part, and both. 
In the extant literature, this approach is also referred to as reparameterization-based methods \cite{ding2023parameter,luo2023towards}. 
\begin{itemize}
  \item \emph{Weight Part}: In this approach, the weight component of the pre-trained model is modified to adapt to the target task. This modification can involve scaling the weight matrix, adding a bias term, or applying a non-linear transformation to the weights. By adjusting the weight component, the model can learn task-specific features while retaining the general knowledge acquired during pre-training. LoRA \cite{hu2021lora}, DyLoRA \cite{valipour2022dylora} are examples of weight part tuning techniques. 
  \item \emph{Bias Part}: In this approach, the bias component of the pre-trained model is adjusted to better align with the target task. By modifying the bias term, the model can learn to make task-specific predictions while leveraging the pre-trained knowledge encoded in the weights. This approach is particularly useful when the target task requires a shift in the model's prediction bias. Bitfit \cite{zaken2021bitfit}, AdapterBias \cite{fu2022adapterbias} are two examples of bias part tuning techniques.
  \item \emph{Weight and Bias Parts}: In this approach, both the weight and bias components of the pre-trained model are simultaneously modified to adapt to the target task. By adjusting both components, the model can learn task-specific features and prediction biases while preserving the general knowledge captured during pre-training. This approach offers a more flexible and comprehensive adaptation strategy that can be tailored to a wide range of downstream tasks. Scale and Shift the deep Features (SSF) \cite{lian2022scaling} is an example of a technique that adjusts both weight and bias components to fine-tune a pre-trained model.
\end{itemize}

\subsubsection{Remapping Tuning}
Remapping tuning represents a class of methodologies that facilitate the transfer of knowledge from a pre-trained model to a novel downstream model, diverging from conventional approaches of direct fine-tuning or processing of the original model. This approach can be categorized into three distinct forms based on how the pre-trained model is utilized: knowledge distillation, weight remapping, and architecture remapping.
\begin{itemize}
  \item \emph{Knowledge Distillation}: Knowledge distillation \cite{buciluǎ2006model,hinton2015distilling} serves a dual purpose in machine learning: it acts as a regularization technique for downstream models and functions as a crucial model compression method. The core principle of knowledge distillation involves transferring learned representations from a large, pre-trained “teacher” model to a smaller “student” model. This transfer is achieved by training the student model to emulate either the final output or intermediate features of the teacher model. The process typically employs a soft target distribution for each instance, facilitating a more nuanced transfer of knowledge. By compelling the downstream model to mimic the pre-trained model's behavior, knowledge distillation not only regularizes the learning process but also enables the creation of more compact models that retain much of the performance of their larger counterparts. 
  \item \emph{Weight Remapping}: Weight remapping is a method that transfers knowledge from a teacher neural network to a student network by directly adjusting the student's weights. Unlike traditional knowledge distillation, which uses the teacher's outputs to train the student, weight remapping aims to make the student behave identically to the teacher for all inputs. Let $f(x; \theta)$ represent the teacher network with input $x$ and parameters $\theta$, and $g(x; \theta')$ represent the student network. The goal is to find new parameters $\theta'$ for the student such that $f(x; \theta) = g(x; \theta')$ for all inputs $x$. This approach potentially preserves more of the teacher's learned features in the student network. Net2Net \cite{chen2015net2net}, EAS \cite{cai2018efficient} are examples of weight remapping techniques. 
  \item \emph{Architecture Remapping}: Architecture remapping is a method used in neural network design, primarily in weight-sharing neural architecture search (NAS) \cite{liu2018darts}. It involves transferring knowledge about network structure from existing models to find optimal new architectures. Essentially, architecture remapping aims to leverage prior knowledge to efficiently discover high-performing network structures, potentially speeding up the process of developing effective neural networks for specific tasks.
\end{itemize}

\section{Key Advances of Recent Geoscience Foundation Models}
\label{sec:key_advances}

With the applications of foundation models for general purposes, the explorations of foundation models for remote sensing-based geoscience tasks have drawn a lot of attention from the research community. In this subsection, we will review the recent advances in GFMs, including the summary of existing foundation models from the view of large language models (LLMs), large vision models (LVMs), large vision-language models (LVLMs), and foundation-model-based agents. Corresponding LLMs, LVMs, and LVLMs for geoscience tasks are also summarized in Table \ref{tab:llm}, Table \ref{tab:lvm}, and Table \ref{tab:lvlm}, respectively.

\subsection{Large Language Models}
LLMs have emerged as a focal point of research in the domains of natural language processing (NLP) and computer vision. These models, characterized by their extensive transformer-based architectures, are designed to facilitate comprehensive understanding of both visual and textual data. LLMs have demonstrated unprecedented efficacy across a spectrum of language-related tasks, including but not limited to language modeling, text generation, and question answering \cite{radford2018improving,devlin2018bert}.

\subsubsection{Basic LLMs}

The GPT (generative pre-trained transformer) model \cite{radford2018improving}, developed by OpenAI, is a pioneering large language model trained on a vast corpus of web text. It has demonstrated impressive performance across various language modeling and text generation tasks. GPT maximizes the likelihood objective
\begin{equation}
\mathcal{L}(\theta) = \sum_{i=1}^{N} \log p(t_i | t_{<i}; \theta),
\end{equation}
where $t = {t_1, ..., t_N }$ is a sequence of tokens, $k$ denotes the context window length, and $\theta$ represents the network parameters. This approach allows GPT to learn contextual relationships in text, enabling it to generate coherent and contextually appropriate language across a wide range of applications.

The GPT series has evolved significantly since its inception. GPT-2 \cite{radford2019language}, with 1.5 billion parameters, achieved remarkable performance across various language tasks, generating human-like text. GPT-3 \cite{radford2019language} introduced zero-shot learning capabilities, sparking interest in in-context learning. InstructGPT \cite{ouyang2022training}, a key component of ChatGPT, improved control and flexibility by incorporating high-level instructions in natural language. The latest iteration, GPT-4 \cite{achiam2023gpt}, has demonstrated the immense potential of large-scale language models, marking a new era in AI's ability to understand and communicate naturally. This evolution highlights the rapid advancements in language model capabilities, from basic text generation to sophisticated, instruction-following AI systems capable of handling a wide range of language tasks with increasing proficiency.

Various BERT model variants have been developed to enhance its performance and efficiency. RoBERTa \cite{liu2019roberta}, developed by Facebook AI, improved BERT through more iterations and data augmentation techniques such as dynamic masking and noising. ALBERT \cite{lan2019albert} introduced factorized embedding parameterization, cross-layer parameter sharing, and a sentence order prediction task to reduce model size and improve training speed. MacBERT \cite{cui2020revisiting} proposed replacing the [MASK] token with synonyms at mask locations instead of using the [MASK] tag. These variants demonstrate ongoing efforts to optimize BERT's architecture, training process, and performance across various natural language processing tasks, showcasing the model's adaptability and potential for further improvements in language understanding and generation.

Beyond GPT and BERT, other significant large language models have emerged. T5 \cite{raffel2020exploring} introduced a unified architecture for various NLP tasks, employing task-agnostic pre-training on a diverse text corpus to create a general-purpose language model. Meanwhile, chain-of-thought (CoT) prompting \cite{wei2022chain} developed a technique enabling intermediate reasoning in large language models. These advancements demonstrate the ongoing evolution in language model capabilities, with T5 aiming for versatility across tasks and CoT enhancing models' ability to perform complex reasoning. Together, these developments showcase the field's progress towards more flexible, task-adaptable, and reasoning-capable language models.

\subsubsection{Advanced LLMs for Geoscience} 
\begin{figure*}
  \centering
  \includegraphics[width=0.95\textwidth]{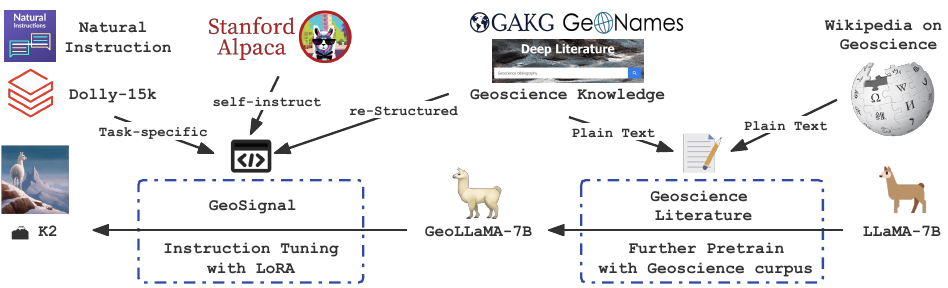}
  \caption{The framework of K2 \cite{deng2023learning}.}
  \label{fig:k2}
\end{figure*}

Table \ref{tab:llm} summarizes the advanced LLMs for geoscience, including the data type, pre-training, adaptation, applications, and key features. 
Several LLMs for geoscience have been developed based on BERT \cite{devlin2018bert}, SciBERT \cite{beltagy2019scibert}, and LLaMA \cite{touvron2023llama}. 
For example, Denli \emph{et al.} \cite{denli2021geoscience} retrained a pre-trained BERT model with our 20M internal geoscientific records, referring to GeoBERT. They fine-tuned the GeoBERT model for several tasks including geoscience question answering and query-based summarization. 
The work \cite{ramachandran2022language} introduced BERT-E, an Earth science-specific language model built using transfer learning from SciBERT \cite{beltagy2019scibert}, which shows improved performance in keyword classification and highlights the need for robust benchmarks in evaluating such domain-specific language models, with potential applications across various scientific fields. 
The authors in \cite{deng2023learning} introduced K2, the first LLM specifically for geoscience, along with a new instruction tuning dataset, GeoSignal, and a benchmark, GeoBench, to adapt and evaluate LLMs in geoscience, demonstrating its effectiveness in understanding and utilizing geoscience knowledge with open-source data and model checkpoints. The framework of K2 is shown in Fig. \ref{fig:k2}. 

For Chinese geological texts, the authors in \cite{ma2023cngeoplm} developed a large-scale Chinese geological corpus (GeoCorpus) and a Chinese pretrained language model (CnGeoPLM) for geological domain, demonstrating that CnGeoPLM significantly outperforms the generic BERT model in geological NLP tasks such as entity recognition, relationship extraction, and entity clustering. Wei \et \cite{wei2022geobertsegmenter} introduced GeoBERTSegmenter, a GeoBERT-based Chinese word segmentation model tailored for geological texts, combining BERT, bidirectional recurrent neural network, and conditional random field to effectively address the segmentation of unregistered geological terms and common words, demonstrating superior performance in precision, recall, and F1 score compared to other methods.

Despite general geoscience tasks, LLMs have been used in specific geoscience domains. For example, the authors in \cite{bi2023oceangpt} presented OceanGPT, the first-ever LLM specialized for ocean science, developed with a novel framework, DoInstruct, for generating domain-specific instruction data and evaluated using OceanBench, a new oceanography benchmark, demonstrating enhanced expertise and preliminary embodied intelligence capabilities in ocean science tasks. 

Additionally, the authors in \cite{lawley2022geoscience} described the development and validation of new geoscience-specific language models trained on public geoscientific reports and publications, which outperform generic models in various NLP tasks and demonstrate the potential of using NLP to extract meaningful semantic relationships and latent knowledge from the vast volume of geoscience literature. The authors in \cite{lawley2023applications} described the application of NLP to geoscientific text data to extract meaningful semantic relationships and predict mineral potential, demonstrating how geoscience language models can transform unstructured text into actionable insights to streamline mineral exploration for critical materials. 
The work \cite{lin2023geogalactica} presented the development of GeoGalactica, the largest language model specialized in geoscience, which is trained on a vast geoscience text corpus and fine-tuned with instruction-tuning data, highlighting the model's comprehensive training process and its open-source contribution to the field. 
The authors in  \cite{roberts2023gpt4geo,ji2023evaluating,bhandari2023large,mooney2023towards,mai2023opportunities}  introduced LLMs for geospatial data analysis, which have been applied to various geospatial tasks such as population density estimation, economic livelihood measurement, and geoscience knowledge understanding, demonstrating the potential of LLMs in enhancing geospatial data analysis and decision-making support.

\subsubsection{Applications and Disscussions} 
The applications of advanced LLMs in geoscience, as illustrated in Table \ref{tab:llm}, demonstrate a diverse and specialized range of capabilities. These span from fundamental text processing tasks such as question answering and summarization (\emph{e.g.}, GeoBERT) to complex analytical functions including knowledge discovery and semantic relationship extraction (\emph{e.g.}, BERT-E, GeoGalactica). There's a strong emphasis on geographic information systems (GIS) related tasks (\emph{e.g.}, GPT4GEO, GEOGPT), reflecting the importance of geospatial data analysis in the field. Specialized applications are evident, with models such as CnGeoPLM focusing on Chinese geological NLP tasks and OceanGPT specializing in ocean science. Practical applications extend to socio-economic assessments, as seen with GeoLLM's use in measuring population density and economic livelihoods. Moreover, models including Geo-RAG focus on geoscience knowledge understanding and utilization, indicating a trend towards LLMs capable of complex reasoning and decision-making support in geoscience. This wide spectrum of applications underscores the versatility and growing significance of LLMs in advancing various aspects of geoscientific research and practice. 

In summary, the development of LLMs for geoscience has been an active and rapidly evolving area of research, with numerous models developed and evaluated for a wide array of geoscience tasks. These models, such as GeoBERT, BERT-E, GeoGalactica, and OceanGPT, have demonstrated significant potential in extracting meaningful insights from geoscience text data and substantially improving the performance of geoscience-specific NLP tasks. Applications range from fundamental text analysis to complex geospatial reasoning and knowledge synthesis, including question answering, summarization, semantic relationship extraction, and integration with Geographic Information Systems. Specialized models address domain-specific and language-specific challenges, while some LLMs extend their utility to practical socio-economic assessments. The trend towards geoscience knowledge understanding and utilization, as seen in models such as Geo-RAG, indicates a shift towards LLMs capable of complex reasoning and decision-making support. As these models continue to evolve, they promise to revolutionize geoscientific data processing, analysis, and utilization, potentially serving as sophisticated AI assistants in research and policy-making. However, challenges remain in areas including model interpretability, data quality, and ethical considerations, highlighting the need for ongoing research and development in this field.

\begin{table*}
  \centering
  \caption{Summary of Advanced Large Language Models for Geoscience}
  \label{tab:llm}
  \begin{tabular}{m{3cm}<{\centering}m{3cm}<{\centering}m{1.5cm}<{\centering}m{1.5cm}<{\centering}m{3cm}<{\centering}m{3cm}<{\centering}}
  \toprule
  \textbf{Ref.} 
  & \textbf{Data Type} 
  & \textbf{Pre-training} 
  & \textbf{Adaptation} 
  & \textbf{Applications} 
  & \textbf{Key Features} \\
  \midrule
  GeoBERT \cite{denli2021geoscience}    
  & Geoscientific records   
  & Generative            
  & -       
  & Question answering and summarization 
  & Retrained on geoscientific records    \\
  \rowcolor{light_gray}
  BERT-E \cite{ramachandran2022language}     
  & Earth science-related articles     
  & Generative        
  & -       
  & Keyword classification       
  & Transfer learning from SciBERT \cite{beltagy2019scibert}     \\
  K2 \cite{deng2023learning}     
  & Geo-text, geoscience instructions correlations         
  & -         
  & Model fine-tuning       
  & Geoscience tasks         
  & Geoscience tasks     \\
  \rowcolor{light_gray}
  CnGeoPLM \cite{ma2023cngeoplm}    
  & Chinese geological corpus     
  & Generative
  & -       
  & Geological NLP tasks          
  & A Chinese pre-trained LVM     \\
  GeoBERTSegmenter \cite{wei2022geobertsegmenter}     
  & Unregistered geological terms        
  & Generative           
  & -         
  & Word segmentation          
  & Chinese word segmentation model       \\
  \rowcolor{light_gray}
  OceanGPT \cite{bi2023oceangpt}     
  & Ocean domain instruction data         
  & Generative            
  & -          
  & Oceans science tasks      
  & The first-ever LLM in the ocean domain      \\
  Lawley \et \cite{lawley2022geoscience}     
  & Geo-scientific reports and publications          
  & Generative           
  & -        
  & NLP tasks          
  & Observation-based data analysis \\
  \rowcolor{light_gray}
  Lawley \et \cite{lawley2023applications}     
  & Geoscientific text data         
  & Generative            
  & -          
  & Extract semantic relationships           
  & New knowledge extraction    \\
  GeoGalactica \cite{lin2023geogalactica}   
  & geoscience text corpus (GeoCoprus)         
  & Generative
  & Model fine-tuning          
  & Question answering \& knowledge discovery          
  & 30 billion parameters     \\
  \rowcolor{light_gray}
  GPT4GEO \cite{roberts2023gpt4geo,ji2023evaluating,bhandari2023large,mooney2023towards,mai2023opportunities}   
  & Geo-text, user instructions
  & -
  & -          
  & Geographic information systems (GIS) related test        
  & LLMs evaluation for GIS   \\
  GEOGPT \cite{zhang2023geogpt}   
  & Geospatial data          
  & -
  & -          
  & Geographic information systems (GIS)         
  & -     \\

  \rowcolor{light_gray}
  BB-GeoGPT \cite{zhang2024bb}   
  & Geo-text, user instructions          
  & Hybrid
  & Prompt tuning          
  & Geospatial questions          
  &      \\

  GeoLLM \cite{manvi2023geollm}   
  & OpenStreetMap          
  & -
  & Prompt tuning
  & measurement of population density and economic livelihoods
  & 70\% improvement in performance     \\
  
  \rowcolor{light_gray}
  GeoLLM-Engine \cite{singh2024geollm}   
  & -          
  & -
  & -          
  & a novel environment 
  & evaluating geospatial Copilots\\

  Zwartjes \et \cite{zwartjes2024building}   
  & Seismic data          
  & -
  & -          
  & Seismic processing assistant          
  & -     \\

  \rowcolor{light_gray}
  Deng \et \cite{deng2024geoscience}   
  & Geoscience knowledge data
  & -
  & -          
  & Geoscience knowledge understanding and utilization
  & -     \\

  Geo-RAG \cite{dong2024geo}   
  & Unstructured Geological Documents
  & -
  & Prompt tuning          
  & Geological Knowledge extraction          
  & -     \\

\bottomrule
\end{tabular}
\end{table*}

\subsection{Large Vision Models}

\subsubsection{Basic Large Vision Models}

The evolution of visual models has seen a significant increase in parameter count. Building on the scalability of Vision Transformers (ViT), Zhai \et \cite{zhai2022scaling} expanded the ViT architecture to encompass 1.8 billion parameters. Google pushed this boundary even further, introducing a model \cite{dehghani2023scaling} with an impressive 22 billion parameters, showcasing visual model scaling capabilities comparable to those of large language models. Concurrently, InternImage \cite{wang2023internimage} demonstrated that CNNs could also be scaled into foundation models (FMs), achieving performance gains similar to ViT-based approaches. While these developments primarily focused on static image processing, researchers have also extended natural image FMs to video applications \cite{liu2022video,wang2022bevt,wang2023videomae}. These advancements have broadened the potential applications of large-parameter FMs across a wide spectrum of visual tasks, encompassing both image and video domains.

\begin{figure}
\centering
\includegraphics[width=0.45\textwidth]{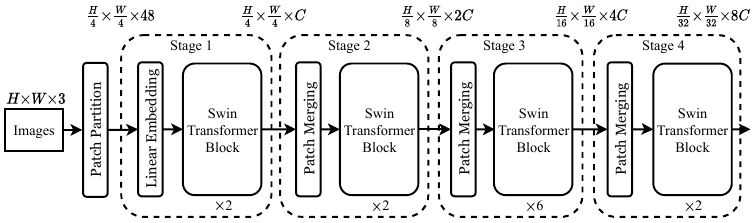}
\caption{The framework of swin transformer \cite{liu2021swin}.}
\label{fig:swin}
\end{figure}

The swin transformer \cite{liu2021swin} introduces a hierarchical architecture reminiscent of CNNs. Unlike traditional vision transformers (ViTs) that maintain constant feature map sizes with a 16x downsampling, swin transformer progressively reduces feature map dimensions as the network deepens. It employs a multi-stage approach with 4x, 8x, and 16x downsampling at different layers. The model's structure, illustrated in Fig. \ref{fig:swin} (a), consists of four key components: patch partitioning, linear embedding, patch merging, and swin transformer blocks. Fig. \ref{fig:swin} (b) depicts the architecture of two consecutive swin transformer blocks. This hierarchical design is crucial for enhancing the model's performance in various visual recognition tasks, effectively combining the strengths of both CNNs and transformers.

\begin{figure}
\centering
\includegraphics[width=0.45\textwidth]{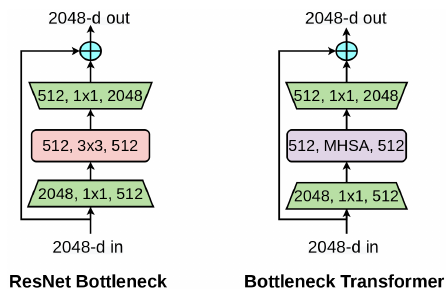}
\caption{Compare the architecture of ResNet bottleneck \cite{he2016deep} and bottleneck.}
\label{fig:bot}
\end{figure}

BotNet \cite{srinivas2021bottleneck} presents a straightforward yet powerful backbone for visual representation, integrating self-attention mechanisms into various computer vision tasks including image classification, object detection, and instance segmentation. The core of BotNet is its Bottleneck Transformer blocks, as illustrated in Fig. \ref{fig:bot}. For a 2048-dimensional input, the traditional ResNet bottleneck employs convolutional operations of $1 \times 1 \times 512$, $3 \times 3 \times 512$, and $1 \times 1 \times 2048$, while maintaining the skip connection. BotNet modifies this structure by replacing the middle $3 \times 3$ convolution with multi-head self-attention (MHSA), which forms the key innovation of the Bottleneck Transformer. MHSA enables the model to capture diverse features and patterns within the input data more effectively. Furthermore, BotNet substitutes the spatial convolutions in ResNet's final three bottleneck blocks with global self-attention. This architectural change significantly enhances performance in instance segmentation and object detection tasks while simultaneously reducing parameter count and computational latency.

The segment anything model (SAM) \cite{kirillov2023segment}, introduced by Meta AI Research, represents a significant advancement in computer vision and image segmentation. SAM is designed as a promptable segmentation system, capable of generating high-quality object masks for any image based on various types of prompts, including points, boxes, or text descriptions. Its architecture combines a powerful image encoder with a prompt encoder and a lightweight mask decoder, enabling it to perform zero-shot segmentation tasks with remarkable flexibility. SAM's training process involved a dataset of over 1 billion masks, contributing to its ability to generalize across diverse objects and scenes. This model's versatility allows it to adapt to a wide range of segmentation tasks without task-specific fine-tuning, making it a valuable tool for numerous applications in image analysis, object detection, and interactive image editing. SAM's introduction has opened new possibilities in the field of computer vision, potentially transforming how we approach image understanding and manipulation tasks.

\begin{figure*}
  \centering
  \includegraphics[width=0.9\textwidth]{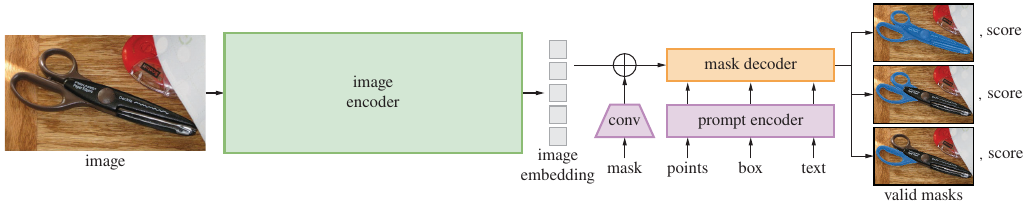}
  \caption{The Segment Anything Model (SAM) architecture \cite{kirillov2023segment}.}
  \label{fig:sam}
\end{figure*}

\subsubsection{Advanced Large Vision Models for Geoscience}
LVMs have been widely adopted in remote sensing and geoscience for various tasks, including remote sensing image analysis, satellite image classification, and climate and weather modeling. Most LVMs are based on the transformer architecture, which has proven effective in handling large-scale remote sensing data. Several studies have introduced general-purpose LVMs for remote sensing applications. For example, Sun \et \cite{sun2022ringmo} introduced RingMo, a remote sensing foundation model framework using generative self-supervised learning, trained on two million remote sensing (RS) images to address domain gap and generalization issues. The work \cite{wang2022advancing} proposed LVMs with a rotated varied-size window attention mechanism, demonstrating superior performance in detection, classification, and segmentation tasks. The authors in \cite{cha2023billion} explored the impact of increasing model parameters on performance, showing improved results and data efficiency in tasks such as rotated object detection and semantic segmentation. Yao \et \cite{yao2023ringmo} presented RingMo-Sense, a foundation model for spatiotemporal prediction in remote sensing, while Wang \et \cite{wang2024ringmo} introduced RingMo-lite, a lightweight CNN-Transformer hybrid network optimized for edge devices.

SAM has been adapted for various remote sensing tasks. The work \cite{yan2023ringmo} introduced RingMo-SAM, a foundation model for multimodal remote-sensing image segmentation. Wang \et \cite{wang2023samrs} created SAMRS, a large-scale remote sensing segmentation dataset using SAM. The authors in \cite{chen2024rsprompter} developed RSPrompter, an automated instance segmentation approach based on SAM. Ma \et \cite{ma2023sam} proposed a novel framework utilizing SAM for semantic segmentation in remote sensing. Sultan \et \cite{sultan2023geosam} introduced Geographical SAM (GeoSAM), an advanced framework for segmenting mobility infrastructure.

Specialized LVMs have been developed for specific remote sensing tasks. For satellite image classification, the work \cite{mai2023csp} introduced CSP, a contrastive pre-trained model, while the authors in \cite{jakubik2023foundation} presented Prithvi, a transformer-based geospatial foundational model pre-trained on over 1TB of multispectral satellite imagery. In weather and climate modeling, Nguyen \et \cite{nguyen2023climax} introduced ClimaX, a flexible deep learning model, and the authors in \cite{man2023w} proposed W-MAE, a novel weather and climate model. Other notable models in this domain include FourCastNet \cite{kurth2023fourcastnet}, GraphCast \cite{lam2022graphcast}, Fengwu \cite{chen2023fengwu}, and Pangu-Weather \cite{bi2023accurate}. For Earth image analysis, Guo \et \cite{guo2023skysense} developed SkySense, a large vision model demonstrating potential in various analysis tasks.

Recent research has also focused on specialized applications of LVMs. Wang \et \cite{wang2024rsbuilding} proposed RSBuilding, a foundation model-based approach for unified building extraction and change detection. The work \cite{sun2024enhancing} explored the use of SAM for crop mapping from satellite imagery. The authors in al. \cite{julka2023knowledge} demonstrated the effectiveness of SAM for knowledge distillation in planetary science, specifically for mapping Martian skylights.

Multi-modal and specialized architectures have also emerged. The authors in \cite{pan4876040vision} introduced VSGNet, a novel multi-modal fusion framework for remote sensing semantic segmentation. Fuller \et \cite{fuller2022satvit} proposed SatViT, a vision transformer pretrained on 1.3 million satellite images using masked autoencoding. The work \cite{li2024s2mae} developed Spatial-SpectralMAE (S2MAE), a specialized pre-trained architecture for spectral remote sensing imagery.

\subsubsection{Applications and Discussions}
Table \ref{tab:lvm} provides an overview of advanced large vision models applied to geoscience, showcasing a diverse range of applications primarily focused on Earth observation and remote sensing tasks. The applications span from fundamental image analysis to complex environmental monitoring and prediction. Key areas include remote sensing image interpretation, segmentation, and classification; spatiotemporal predictions; infrastructure and building analysis; climate and weather forecasting; Earth and planetary image analysis; and various specialized tasks such as glaciological studies, multispectral object detection, and local climate zone classification. These models utilize different types of geospatial data, including satellite imagery, multispectral data, and other remote sensing inputs, to address a wide spectrum of geoscience challenges. The variety of applications demonstrates the versatility and potential of large vision models in advancing our understanding and analysis of Earth systems across multiple scales and domains. 

LVMs have emerged as transformative technologies in geoscience and remote sensing, demonstrating remarkable capabilities across tasks such as image analysis, classification, segmentation, and climate modeling. The adoption of transformer-based architectures has enabled these models to effectively handle the complexity and scale of geospatial data, leading to significant improvements in performance and efficiency. Models such as RingMo, Prithvi, ClimaX, and adaptations of the SAM have shown the potential to address domain-specific challenges in remote sensing, climate science, and Earth observation. As the field advances, key trends such as increased multimodal integration, enhanced spatial and temporal reasoning, improved explainability, and a focus on efficiency and ethical considerations are likely to shape future developments. These advancements promise to unlock new insights into our planet, enhance our ability to monitor and respond to environmental changes, and revolutionize how geoscientists, policymakers, and environmental managers analyze and understand complex geospatial data.

\begin{table*}
  \centering
  \caption{Summary of Advanced Large Vision Models for Geoscience}
  \label{tab:lvm}
  \begin{tabular}{m{3cm}<{\centering}m{3cm}<{\centering}m{1.5cm}<{\centering}m{1.5cm}<{\centering}m{3cm}<{\centering}m{3cm}<{\centering}}
  \toprule
  \textbf{Ref.} 
  & \textbf{Data Type} 
  & \textbf{Pre-training} 
  & \textbf{Adaptation} 
  & \textbf{Applications} 
  & \textbf{Key Features} \\
  \midrule
  RingMo \cite{sun2022ringmo}, Wang \et \cite{wang2022advancing}, Cha \et \cite{cha2023billion} 
  &  Remote sensing images
  &   Generative
  &   Fine-tuning      
  &   Remote sensing image analysis          
  &    -  \\
  \rowcolor{light_gray}
  RingMo-Sense \cite{yao2023ringmo}
  &  Remote sensing images
  &   Generative
  &   Fine-tuning      
  &   Spatiotemporal prediction          
  &   Video Swin Transformer    \\
  RingMo-Lite \cite{wang2024ringmo}
  &  Remote sensing images
  &   Generative
  & -
  &   RS image interpretation
  &   Optimized for edge devices    \\
  \rowcolor{light_gray}
  RingMo-SAM \cite{yan2023ringmo}
  &  Remote sensing images
  &  Generative
  &  Fine-tuning
  &  Remote sensing image segmentation
  &  Effective across various datasets    \\
  SAMRS \cite{wang2023samrs}, RSPrompter \cite{chen2024rsprompter}
  &  Remote sensing images
  &   -
  &  Prompt tuning
  &  Remote sensing image segmentation
  &   SAM    \\
  \rowcolor{light_gray}
  Ma \et \cite{ma2023sam}
  &  Remote sensing images
  &  Generative
  &  Fine-tuning
  &  Remote sensing image segmentation
  &  Segment Anything Model (SAM)    \\
  Geo-SAM \cite{sultan2023geosam}
  & Remote sensing images
  & -
  & Fine-tuning
  & Infrastructure segmentation
  & Segment Anything Model (SAM)    \\
  \rowcolor{light_gray}
  OFA-Net \cite{xiong2024one}
  & Remote sensing images
  & -
  & Fine-tuning
  & Infrastructure segmentation
  & Segment Anything Model (SAM)    \\
  CSP \cite{mai2023csp}
  &  Remote sensing images
  &  Contrastive
  &  Fine-tuning
  &   Satellite image classification
  &   Satellite   \\
  \rowcolor{light_gray}
  Prithvi \cite{jakubik2023foundation}
  & Satellite imagery
  & Generative
  & Fine-tuning
  & Earth observation tasks
  & open source \\
  ClimaX \cite{nguyen2023climax}, W-MAE \cite{man2023w}
  &  Grid-based meteorological data
  &   Generative
  &   Fine-tuning
  &   Climate change prediction
  &   Climate and weather modeling    \\
  \rowcolor{light_gray}
  Fourcastnet \cite{kurth2023fourcastnet}, GraphCast \cite{lam2022graphcast}, Fengwu \cite{chen2023fengwu}, Pangu-Weather \cite{bi2023accurate}
  &  Grid-based meteorological data
  &   Generative
  & -
  &   Weather forecasting
  &   Climate and weather modeling    \\
  SkySense \cite{guo2023skysense}
  &  Multi-modal Remote Sensing Imagery dataset
  &   Contrastive
  &   Fine-tuning
  &   Earth image analysis
  &   State-of-the-art performance    \\
  \rowcolor{light_gray}
  Shankar \et \cite{shankar2023semantic,shankar2023segment}
  &  Sentinel-1 and Sentinel-2 imagery
  &   Contrastive
  &   Fine-tuning
  &   glaciological image analysis
  &   -    \\
  Julka \et \cite{julka2023knowledge}
  &  HiRISE Images
  &   Hybrid
  &   Remapping Tuning
  &   Planetary image analysis
  &   SAM \\

  \rowcolor{light_gray}
  Osco \et \cite{osco2023segment}
  &  Remote sensing images
  &   Hybrid
  &   Fine-tuning
  &   Satellite imagery analysis
  &   SAM\\

  Sun \et \cite{sun2024enhancing}
  &  Satellite imagery
  &   Hybrid
  &   Fine-tuning
  &   Parcel segmentation
  &   SAM    \\

  \rowcolor{light_gray}
  RSbuilding \cite{wang2024rsbuilding}
  &  Satellite imagery
  &   Hybrid
  &   -
  &   Building extraction and change detection
  &   -    \\

  VSGNet \cite{pan4876040vision}
  &  Multi-modal RS images
  &   -
  &   Fine-tuning
  &   RS semantic segmentation
  &   -    \\

  \rowcolor{light_gray}
  Satvit \cite{fuller2022satvit}
  &  Satellite-derived RS images
  &   Contrastive
  &   Fine-tuning
  &   RS images analysis
  &   ViT    \\

  SpectralGPT \cite{hong2024spectralgpt}   
  & Spectral remote sensing images
  & Generative 
  & Fine-tuning          
  & captioning and retrieving remote sensing images          
  & 600 million parameters    \\
  \rowcolor{light_gray}
  Rsmamba \cite{chen2024rsmamba}   
  & remote sensing images
  & -
  & -
  & remote sensing image classification
  & -     \\
  Ren \et \cite{ren2024segment}   
  & overhead imagery
  & -
  & -
  & various overhead imagery tasks
  & SAM evaluation\\

  \rowcolor{light_gray}
  DF4LCZ \cite{wu2024df4lcz} 
  & Google imagery \& multispectral imagery
  & -
  & -
  & local climate zones classification          
  & -     \\

  Gallagher \et \cite{gallagher2024multispectral}   
  & Multispectral imagery
  & -
  & Prompt tuning          
  & multispectral object detection          
  & SAM \\
  \rowcolor{light_gray}
  S2MAE \cite{li2024s2mae}  
  & Spectral RS images          
  & -             
  & -      
  & -           
  & A million training images
  \\
  Feng \et \cite{feng2023self}  
  & Cross-modal data          
  & -             
  & -      
  & Multiple downstream tasks
  & 
  \\

  \bottomrule
\end{tabular}
\end{table*}
  
\subsection{Large Vision-Language Models} 

\begin{figure*}
  \centering
  \includegraphics[width=0.9\textwidth]{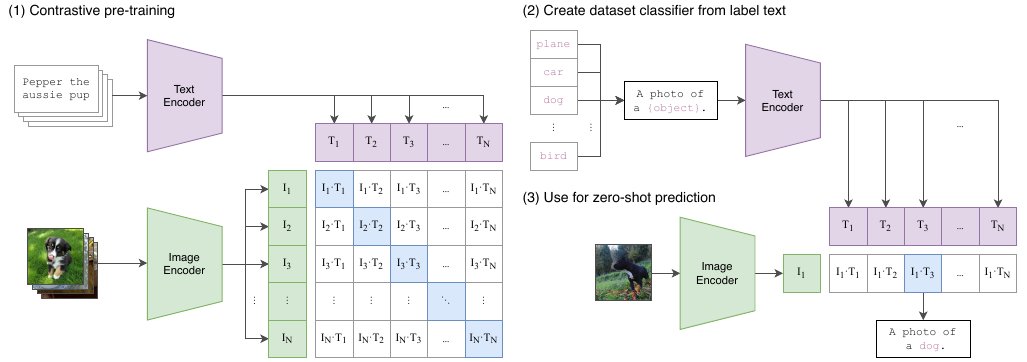}
  \caption{CLIP \cite{radford2021learning}.}
  \label{fig:clip}
\end{figure*}

\subsubsection{Basic Large Vision-Language Models}
In the current landscape, FMs have evolved to encompass more than just deep models; they now focus on utilizing vast amounts of data and computational power to tackle diverse problems. The goal is to use a unified model capable of addressing multiple modalities and tasks. Consequently, there is a growing emphasis on training vision-language models \cite{zhang2024vision}.

CLIP \cite{radford2021learning} is a prominent example that leverages a large-scale collection of image-text pairs from the Internet for contrastive learning, enabling the creation of a unified representation of
multimodal data. 
Inspired by CLIP, various multimodal image FMs have been proposed, differing in their model structures \cite{wang2023image,alayrac2022flamingo}, feature representations \cite{yuan2021florence,wang2022internvideo}, multimodal feature fusion approaches \cite{li2021align}, feature alignment loss functions \cite{yu2022coca}, pre-training methods \cite{li2023blip}, and more. These advancements have significantly improved the performance of multimodal FMs.

In addition, DALL-E \cite{ramesh2022hierarchical} combines the diffusion model with multimodal FMs to generate images from text. SAM \cite{kirillov2023segment} introduces a promptable model with training strategies that enable the segmentation of objects using text, points, and lines as prompts. GPT-4 \cite{achiam2023gpt} exhibits superhuman capabilities on various professional and academic datasets. mPLUG-2 \cite{xu2023mplug} introduces a multimodule composition network, including text, image, and video. Metatransformer \cite{zhang2023meta} proposed a unified framework performing learning across 12 modalities with unpaired data (e.g., natural language, 2-D images, 3-D point clouds, audio, video, time series, and tabular data).

\subsubsection{Advanced Large Vision-Language Models for Geoscience}

LVLMs have emerged as powerful tools in remote sensing and geoscience, particularly for visual question answering and image understanding tasks. Chappuis \et \cite{chappuis2022language} pioneered the use of Remote Sensing Visual Question Answering (RSVQA), leveraging advancements in natural language processing and computer vision to analyze satellite images. This work was further developed by the work \cite{hackel2023lit}, who introduced LiT-4-RSVQA, a lightweight transformer-based architecture for efficient and accurate visual question answering in remote sensing. However, the authors in \cite{chappuis2023curse} later identified the need to address language biases in RSVQA, revealing more severe biases in remote sensing compared to standard VQA due to dataset specifics.

In the realm of remote sensing image understanding, several notable developments have occurred. Kuckreja \et \cite{kuckreja2023geochat} introduced GeoChat, the first remote sensing VLM with multitask conversational capabilities. The authors in \cite{zhang2024earthgpt} and the work \cite{zhan2024skyeyegpt} presented EarthGPT and SkyeyeGPT respectively, both LVLMs tailored for Earth observation and remote sensing tasks. To address the lack of comprehensive datasets, Hu \et \cite{hu2023rsgpt} developed a high-quality remote sensing image captioning dataset (RSICap) and a benchmark evaluation dataset (RSIEval).

Evaluation and benchmarking of LVLMs in remote sensing have been crucial for understanding their capabilities and limitations. The authors in \cite{zhang2024good} conducted a comprehensive benchmark focusing on scene understanding, localization, counting, and change detection tasks, revealing that while these models excel in tasks such as location understanding and captioning, their spatial reasoning limitations affect performance in object localization and counting. Li \et \cite{li2024vrsbench} introduced VRSBench, a comprehensive vision-language benchmark for remote sensing image understanding, featuring a large dataset of verified captions, object references, and question-answer pairs.

Recent research has focused on developing more advanced LVLM architectures and techniques specifically tailored for remote sensing applications. Liu \et \cite{liu2024remoteclip} presented RemoteCLIP, a vision-language foundation model that leverages data scaling and incorporates UAV imagery. The work \cite{djoufack2022clip} developed CLIP-RS, a cross-modal remote sensing image retrieval platform. The authors in \cite{pang2024h2rsvlm} introduced H2RSVLM, a model trained on a large dataset for enhancing self-awareness in remote sensing image understanding tasks.

Further advancements include the work \cite{al2023vision}, who explored the use of CLIP/Open-CLIP variants for zero-shot classification in remote sensing, and He \et \cite{he2023foundation}, who proposed a foundation model adaptation (FMA) framework for multimodal remote sensing data classification. The authors in \cite{feng2023self} developed a remote sensing cross-modal foundation model that uses multi-domain representation and cross-domain fusion to extract generalized features from massive multi-modal data.

\subsubsection{Applications and Disscussion}
LVLMs have found diverse applications in remote sensing, including visual question answering \cite{chappuis2022language}, scene classification \cite{li2023rs}, object detection, change detection, image captioning, and multi-modal data fusion \cite{feng2023self}. These applications have the potential to revolutionize fields such as urban planning, environmental monitoring, and disaster response. However, the advancement of LVLMs in remote sensing also presents challenges. While models such as GeoChat \cite{kuckreja2023geochat} and lightweight architectures such as LiT-4-RSVQA \cite{hackel2023lit} improve accessibility and efficiency, issues of bias \cite{chappuis2023curse}, interpretability, and ethical use remain critical concerns. The development of high-quality datasets \cite{hu2023rsgpt} and benchmarks \cite{li2024vrsbench} is crucial for progress, as is the integration of domain-specific knowledge. As LVLMs continue to evolve, balancing their immense potential with careful consideration of fairness, privacy, and security will be essential to their successful application in Earth observation and geospatial analysis.

The rapid advancement of LVLMs in geoscience and remote sensing presents both opportunities and challenges for the field. While models such as GeoChat \cite{kuckreja2023geochat} and RemoteCLIP \cite{liu2024remoteclip} demonstrate the potential to democratize access to Earth observation data and improve performance across various tasks, they also raise important concerns. The work by Chappuis \et \cite{chappuis2023curse} on language biases in RSVQA highlights the need for careful consideration of fairness and bias mitigation, especially given the global nature of satellite imagery. As these models become more complex and widely used, ensuring interpretability and trustworthiness becomes crucial, particularly for applications in critical areas such as disaster response or environmental monitoring. The development of high-quality datasets and benchmarks, as seen in the work of Hu \et \cite{hu2023rsgpt} and Li \et \cite{li2024vrsbench}, is essential for advancing the field, but creating large-scale, diverse, and accurately labeled datasets remains a challenge. Furthermore, as LVLMs become more powerful in analyzing high-resolution satellite imagery, addressing privacy and security concerns becomes increasingly important. Future research should focus on integrating domain-specific knowledge, improving model generalization, and developing ethical guidelines for the responsible use of these technologies in Earth observation and geospatial analysis.

\begin{table*}
  \centering
  \caption{Summary of Advanced Large Vision-Language Models for Geoscience}
  \label{tab:lvlm}
  \begin{tabular}{m{3cm}<{\centering}m{3cm}<{\centering}m{1.5cm}<{\centering}m{1.5cm}<{\centering}m{3cm}<{\centering}m{3cm}<{\centering}}
  \toprule
  \textbf{Ref.} 
  & \textbf{Data Type} 
  & \textbf{Pre-training} 
  & \textbf{Adaptation} 
  & \textbf{Applications} 
  & \textbf{Key Features} \\
  \midrule  
  RSVQA\cite{chappuis2022language}    
  &  Images and text         
  &  Generative           
  &  Model fine-tuning          
  &  Satellite image analysis           
  &  Satellite image analysis      \\
  \rowcolor{light_gray}
  LiT-4-RSVQA \cite{hackel2023lit}    
  & Images and text          
  & Generative             
  & Model fine-tuning      
  & Vision question answering            
  & Lightweight        \\
  Chappuis \et \cite{chappuis2023curse}  
  &  -       
  &  -           
  &  -    
  &  -           
  &  Exploration of language biases in RSVQA      \\
  \rowcolor{light_gray}
  GeoChat \cite{kuckreja2023geochat}   
  &   Multimodal geoscience data       
  &  -             
  &  Prompt tuning       
  &  Geospatial data analysis            
  & Robust zero-shot performance        \\
  EarthGPT \cite{zhang2024earthgpt}
  &  Image-text pairs        
  &  -
  &  Instruction tuning     
  & Universal RS image comprehension           
  &  Open-set reasoning tasks.     \\
  \rowcolor{light_gray}
  SkyeyeGPT \cite{zhan2024skyeyegpt,wang2024skyscript}
  & RS instruction-following dataset
  & -
  & Instruction tuning
  & Image-level and region-level tasks
  & Comparable results with GPT-4V  \\
  Mall \et \cite{mall2023remote}
  &  Multimodal geospatial data
  &  Generative
  &  Fine-tuning
  &  Geospatial data analysis
  &  Satellite images \\
  \rowcolor{light_gray}
  RS-CLIP \cite{li2023rs}
  & Image–text datasets        
  & Contrastive            
  & Fine-tuning     
  & Remote sensing scene classification           
  & Zero-shot remote sensing scene classification      \\
  Zhang \et \cite{zhang2024good}  
  & Images and text          
  & -             
  & -      
  & -           
  & Evaluation of LVLMs in Earth observation tasks        \\
  \rowcolor{light_gray}
  Datasets\cite{hu2023rsgpt,yuan2024chatearthnet,xu2024rs} 
  &  Image-text datasets         
  &  -            
  &  -          
  &  Remote sensing           
  &  RSICap and RSIEval datasets for VLMs      \\

  RemoteCLIP \cite{liu2024remoteclip}  
  & Unified image-caption data          
  & Hybrid             
  & Fine-tuning      
  & Various downstream tasks           
  & First general-purpose VLM for RS   \\

  \rowcolor{light_gray}
  CLIP-RS \cite{djoufack2022clip}
  & RS images and text          
  & Contrastive 
  & Fine-tuning   
  & Cross-modal RS image retrieval         
  & CLIP  \\

  H2RSVLM \cite{pang2024h2rsvlm}  
  & RS images with captions          
  & -             
  & -      
  & Large scale captions for RS images          
  & 1.4 million image-caption pairs        \\

  \rowcolor{light_gray}
  Al Rahhal \et \cite{al2023vision}  
  & RS images with captions          
  & -             
  & Prompt tuning
  & Zero-shot RS classification          
  & CLIP/Open-CLIP        \\

  VRSBench \cite{li2024vrsbench}  
  & Images, captions, question-answer pairs          
  & -             
  & -      
  & Remote sensing images           
  & A new benchmark        \\

  \rowcolor{light_gray}
  He \et \cite{he2023foundation}  
  & Multimodal RS datasets          
  & -             
  & -      
  & Multimodal RS data classification           
  & Foundation model adaptation (FMA) framework
  \\

  Prompt-RSVQA \cite{chappuis2022prompt}  
  & Visual images with questions and answers         
  & Hybrid           
  & Prompt tuning
  & VQA in remote sensing           
  & -
  \\

  \rowcolor{light_gray}
  RS-CapRet \cite{silva2024large}   
  & Remote sensing images
  & Contrastive 
  & Fine-tuning          
  & Captioning and retrieving remote sensing images          
  & -     \\

  Text2seg \cite{zhang2023text2seg}   
  & Remote sensing images
  & -
  & Prompt Tuning
  & Remote sensing image segmentation          
  & -     \\
  \bottomrule
\end{tabular}
\end{table*}

\subsection{Foundation-Model-based Agents}
The concept of "agents" has a rich and multifaceted history, traversing diverse fields from philosophy \cite{wooldridge1995intelligent} and cognitive science to artificial intelligence \cite{lior2019ai}. Originally rooted in discussions of agency and autonomy in human behavior, the term has evolved to encompass a wide range of entities capable of perceiving their environment and taking actions to achieve specific goals. 
For more detials about general foundation-model-based agents, please refer to \cite{xi2023rise,wang2024survey}.

\subsubsection{Basic Foundation-Model-based Agents}
\begin{figure*}
  \centering
  \includegraphics[width=0.85\textwidth]{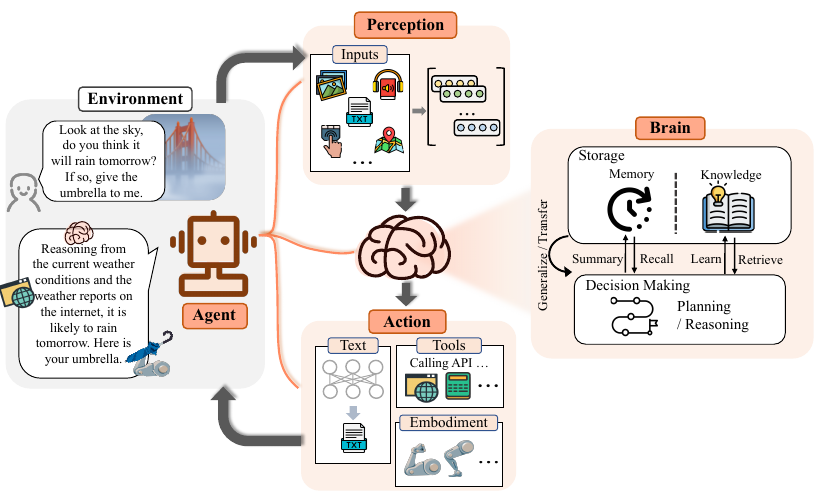}
  \caption{Conceptual framework of LLM-based agent \cite{xi2023rise}.}
  \label{fig:llm-agent}
\end{figure*}

As shown in Fig. \ref{fig:llm-agent}, a typical large language model (LLM)-based agent consists of three main components: brain, perception, and action. 
\begin{itemize}
  \item \textbf{Brain}: The brain module is responsible for basic tasks, including memorizing, thinking, and decision-making.
  \item \textbf{Perception}: The perception module processes the input data, such as text, images, or other modalities, and converts it into a format that the brain can understand.
  \item \textbf{Action}: The the action module carries out the execution using tools and influences the surroundings.
\end{itemize}
When asked about rain, the agent's perception module interprets the question, while the brain module reasons based on current weather conditions and online weather reports. The action module then responds and potentially offers an umbrella, with this process repeating to allow continuous interaction and feedback between the agent and its environment.

\subsubsection{Advanced Foundation-Model-based Agents for Geoscience}
\begin{figure}
  \centering
  \includegraphics[width=0.4\textwidth]{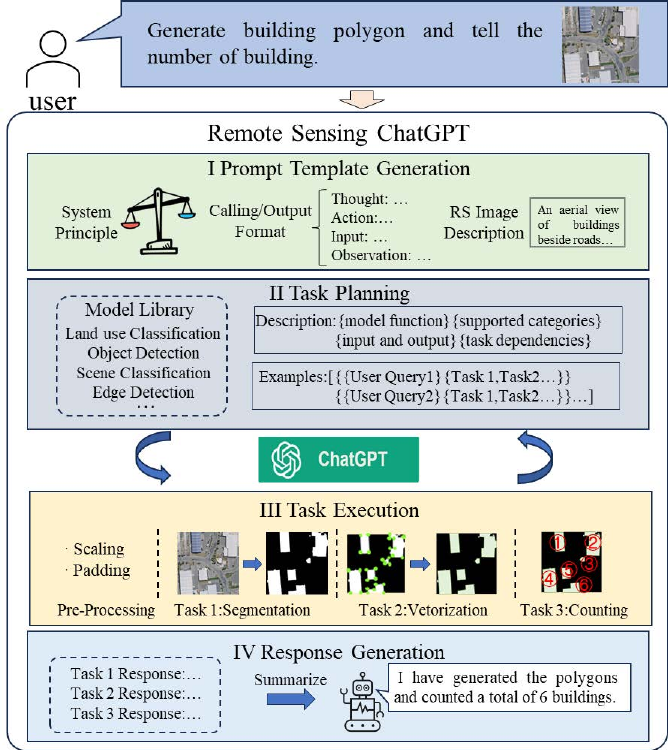}
  \caption{The workflow of Remote Sensing ChatGPT \cite{guo2024remote}.}
  \label{fig:rs-chatgpt}
\end{figure}

There are only several studies on foundation-model-based agents for geoscience, particularly remote sensing (RS). 
For example, RS-ChatGPT \cite{guo2024remote} includes the process of prompt template generation, task planning, task execution, and response generation, as shown in Fig. \ref{fig:rs-chatgpt}. 
RS-ChatGPT operates through a three-step workflow: prompt template generation, task planning, and task execution with response generation. Initially, it creates a prompt template that guides ChatGPT in processing remote sensing tasks, incorporating strict file handling, tool-based execution, and image captioning. Next, it plans tasks by generating detailed descriptions of various remote sensing functions, which are integrated into the prompt template to enhance ChatGPT's understanding. Finally, ChatGPT executes the planned tasks by selecting and applying appropriate tools to preprocess remote sensing images, iteratively feeding results back as new observations until a final response is generated based on all executed tasks. This workflow enables RS-ChatGPT to effectively analyze remote sensing data despite ChatGPT's inability to directly access images, with the potential for further expansion and improvement of its capabilities.

Similarly, the authors in \cite{liu2024change} proposed an interactive Change-Agent system, innovating RS image change interpretation by combining a multilevel change interpretation (MCI) model with a large language model. This integration allows for comprehensive analysis of Earth's surface changes, including detection, captioning, object counting, and cause analysis. The MCI model, trained on a new dataset, uses pixel-level and semantic-level processing enhanced by a novel interaction layer. Experiments show top performance in both change detection and description. 
Xu \et \cite{xu2024rsagent} explored RS-Agent, which is an advanced RS AI that combines a large language model, specialized image processing tools, and expert knowledge. This integration allows it to understand complex queries, perform multi-step analyzes, and provide professional insights in RS tasks. Tested on various datasets, RS-Agent shows superior performance in scene classification, visual question answering, and object counting, demonstrating its ability to handle complex RS applications more effectively than previous models.

\subsubsection{Applications and Discussions}
The development of foundation-model-based agents for geoscience, particularly RS, has the potential to revolutionize the field by enabling intelligent systems capable of understanding and processing complex geospatial data. RS-ChatGPT \cite{guo2024remote}, Change-Agent \cite{liu2024change}, and RS-Agent \cite{xu2024rsagent} represent a significant step forward in this direction, demonstrating the feasibility of using LLMs to perform a wide range of RS tasks, including scene classification, land use classification, object detection, image captioning, edge detection, polygonization, object counting, and change detection. 
In the future, foundation-model-based agents are expected to play an increasingly important role in geoscience research, enabling more efficient data analysis, decision-making, and knowledge discovery.

% %%%%%%%%%%%%%%%%%%%%%%%%%%%%%%%%%%%%%%%%%%%%%%%%%%%%%%%%%%%%%

\section{Challenges and Future Trends} % (fold)
\label{sec:challenges_and_future_trends}

\begin{figure*}
  \centering
  \includegraphics[width=0.95\textwidth]{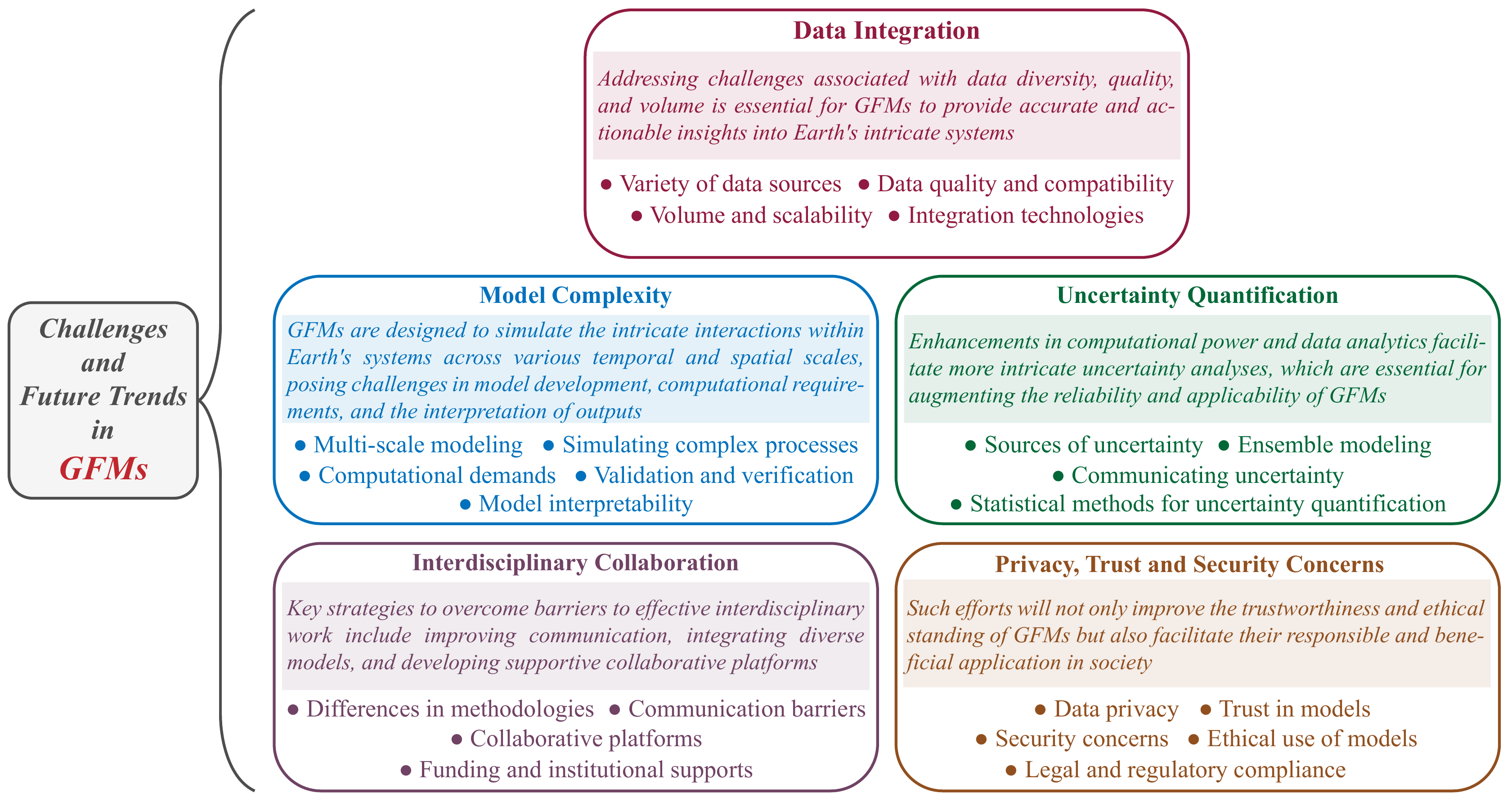}
  \caption{Challenges and Future Trends in GFMs.}
  \label{fig:future}
\end{figure*}

In this section, we discuss the challenges and future trends in GFMs, including data integration, model complexity, uncertainty quantification, interdisciplinary collaboration, and privacy, trust and security concerns. The challenges and future trends in GFMs are summarized in Fig. \ref{fig:future}.

\subsection{Data Integration} 
Effective data integration is pivotal for the development of reliable GFMs. Addressing challenges associated with data diversity, quality, and volume is essential for GFMs to provide accurate and actionable insights into Earth's intricate systems. In geoscience, the integration of diverse datasets into a coherent, unified model that accurately represents these systems poses a significant challenge. This process is further complicated by the heterogeneity of data sources, variable data quality, and the substantial volumes of data produced by modern sensing technologies.

\subsubsection{Variety of Data Sources} 
Geoscientific data originates from a variety of sources, including satellite observations, geological surveys, atmospheric sensors, and oceanographic instruments \cite{conrad2011review,sun2019geospatial,mansfield2020predicting,yu2021deep,casagli2023landslide,runge2023causal}. These sources produce data in diverse formats, scales, and resolutions, complicating their integration into a unified model. For instance, satellite data may offer extensive spatial coverage but lack the temporal resolution needed for real-time monitoring. In contrast, ground-based sensors can provide high-frequency data that is, however, geographically restricted \cite{manfreda2018use}.

\subsubsection{Data Quality and Compatibility} 
The quality of geoscience data varies significantly depending on the source and collection method \cite{karpatne2018machine}. Common issues such as data inaccuracies, missing values, and measurement errors can introduce significant biases into model outputs if not adequately addressed. Moreover, ensuring data compatibility across different scales and formats necessitates sophisticated preprocessing techniques, such as normalization, error correction, and interpolation, to guarantee meaningful contributions from all data inputs to the models.

\subsubsection{Volume and Scalability}
Foundation models, trained on extensive datasets, typically possess billions of parameters \cite{bommasani2021opportunities}. These models demand substantial computational resources for training, requiring thousands of GPUs or TPUs and consuming considerable energy. Additionally, the rapid increase in data produced by digital technologies poses a significant challenge. Managing and processing these large datasets necessitates advanced computational resources and sophisticated data management systems. Scalability is essential, as models must efficiently process increasing volumes of data without compromising processing speed or accuracy \cite{al2015efficient}.

\subsubsection{Integration Techniques} 
Addressing these challenges often entails employing advanced data integration techniques, including sophisticated machine learning algorithms that automate data cleaning, integration, and analysis \cite{roh2019survey}. Geographic Information Systems (GIS) and other spatial analysis tools are crucial in aligning disparate data types across spatial dimensions \cite{sun2019geospatial}. Additionally, data assimilation techniques, traditionally used in meteorology and oceanography to systematically combine model predictions with observational data, are being adapted for broader geoscience applications to improve model accuracy and reliability.

\subsection{Model Complexity}

Addressing the complexities inherent in GFMs is essential to enhancing their functional capabilities. Critical areas for ongoing research and development encompass streamlining models to retain essential details effectively, advancing computational techniques, and enhancing stakeholder communication. GFMs are designed to simulate the intricate interactions within Earth's systems across various temporal and spatial scales, a complexity that poses substantial challenges in model development, computational requirements, and the interpretation of outputs.

\subsubsection{Multi-Scale Modeling}
GFMs require the capacity to operate effectively across a range of scales. Specifically, global climate models must integrate data and processes at the planetary scale while concurrently delivering accurate predictions at regional and local levels \cite{gettelman2022future}. This multi-scale requirement complicates the modeling process, as each scale may necessitate unique modeling approaches and distinct data resolutions to accurately reflect the varying dynamics and interactions \cite{tsai2021calibration}.

\subsubsection{Simulating Complex Processes}
Earth systems are composed of a wide range of dynamic processes, including atmospheric dynamics, hydrological cycles, geological transformations, and ecological interactions \cite{reichstein2019deep,steffen2020emergence}. Each process exhibits significant inherent complexity, yet primary challenges stem from accurately modeling the interactions among these diverse elements. For instance, effectively predicting the impacts of climate change on river basins requires a deep understanding not only of atmospheric physics and hydrology but also of relevant human activities \cite{wu2019impact}. This necessitates an integrated approach that encompasses multiple scientific disciplines to ensure that models can accurately reflect the multifaceted nature of earth systems.

\subsubsection{Computational Demands}
The complexity inherent in GFMs requires extensive computational resources. These models typically incorporate advanced numerical methods, utilize high-resolution data, and necessitate prolonged simulation periods, all demanding substantial computational power. As a result, high-performance computing (HPC) systems are often indispensable, leading to considerable expenses and requiring specialized technical expertise \cite{reed2022reinventing}. This reliance on HPC underscores the critical need for optimization and efficiency improvements in model algorithms to manage costs and accessibility issues effectively.

\subsubsection{Validation and Verification}
Unlike conventional geoscience AI models, which are designed for specific tasks and validated within predefined use case parameters, GFMs are engineered to perform novel tasks as specified by end-users without additional training \cite{bommasani2021opportunities}. This capability considerably complicates predicting potential failure modes in such versatile systems. Effective testing of GFMs requires accumulating illustrative cases over extended periods to statistically capture rare events \cite{bi2023accurate,zhang2023skilful}. Furthermore, as these models become more complex, the challenge of ensuring accurate predictions intensifies. Validating such models often necessitates extensive field data, which can be difficult and costly to obtain. Additionally, with increasing complexity, it becomes progressively challenging to identify the components responsible for inaccuracies in the outputs.

\subsubsection{Model Interpretability}
Interpretability remains an active field of research, continuously evolving with the development of new techniques. The degree of interpretability varies, as no model is entirely transparent or utterly opaque. Researchers are consistently working on methods to demystify the "black box" nature of foundation models while recognizing their inherent limitations \cite{jiang2024interpretable}. Foundation models, such as ChatGPT-4 and Claude-3.5, present interpretative challenges due to their substantial scale and complexity \cite{chen2023fengwu,xie2024wildfiregpt}. Possessing billions of parameters, these models make it challenging to elucidate the reasons behind their specific outputs. Consequently, the complexity of GFMs can hinder stakeholders, including policymakers and the public, from fully comprehending and utilizing the insights these models provide. This challenge may prevent the effective integration of scientific discoveries into decision-making processes crucial for environmental management and policy formulation. 

\subsection{Uncertainty Quantification}

Effective uncertainty quantification is crucial for enhancing the credibility and utility of GFMs. Ongoing advancements in statistical modeling, computational techniques, and stakeholder communication are vital for managing the inherent uncertainties in modeling Earth systems \cite{doyle2019communicating,rounsevell2021identifying}. Accurate quantification of uncertainty in GFMs is essential for delivering reliable predictions and informing risk management and decision-making processes. However, the inherent unpredictability of natural systems, combined with limitations in data and modeling techniques, presents significant challenges \cite{reichstein2019deep}. Ongoing research in uncertainty quantification of GFMs is dedicated to devising more sophisticated methods that can incorporate various types of uncertainties, thereby offering more accurate and comprehensive risk assessments. Enhancements in computational power and data analytics facilitate more intricate uncertainty analyses, which are essential for augmenting the reliability and applicability of GFMs.

\subsubsection{Sources of Uncertainty}
Uncertainties in GFMs stem from multiple sources, such as incomplete or inaccurate data, simplified assumptions in the modeling, and the inherent randomness of natural processes. Specifically, the prediction of earthquake occurrences illustrates these complexities well. Uncertainties here often arise from an incomplete understanding of subsurface geological conditions combined with the inherently chaotic nature of seismic processes \cite{hulbert2019similarity,kuang2021real}. This multifaceted origin of uncertainties necessitates sophisticated modeling techniques to enhance accuracy and reliability in predictions.

\subsubsection{Statistical Methods for Uncertainty Quantification}
Geoscientists utilize a variety of statistical methods to tackle uncertainties in their models. Probabilistic models, notably Bayesian networks, are employed to embed uncertainty directly within the model predictions \cite{gawlikowski2023survey}. These models yield a spectrum of possible outcomes rather than a singular deterministic result, thus providing a more nuanced understanding of potential scenarios. Additionally, Monte Carlo simulations represent another widely used approach \cite{abdar2021review}. This technique involves repeatedly running the model with varied inputs to produce a distribution of potential outcomes, thereby capturing the variability and uncertainty inherent in natural systems.

\subsubsection{Ensemble Modeling}
Ensemble modeling has become a fundamental technique in fields such as climate science and hydrology \cite{zounemat2021ensemble}. This method involves running multiple distinct models or a single model under varying parameter settings to explore a wider array of scenarios. Such a comprehensive approach enables scientists to better capture the uncertainties inherent in system responses. Additionally, ensemble modeling facilitates the identification of predictions that are consistent across different models and those that are particularly sensitive to specific assumptions. This dual capability enhances the reliability of model outputs and aids in refining model assumptions based on comparative robustness.

\subsubsection{Communicating Uncertainty}
A significant challenge in the field of geoscience involves effectively communicating the uncertainties associated with model predictions to non-specialist stakeholders. It is imperative for decision-makers to comprehend the level of confidence in model outputs to facilitate informed decision-making. This necessitates not only the transparent reporting of uncertainties but also proactive education and engagement with the users of model data. Such interactions help ensure that stakeholders fully understand the implications of uncertainty within the context of their specific decision-making processes.

\subsection{Interdisciplinary Collaboration}
The development of comprehensive GFMs requires collaboration across multiple scientific disciplines, including climatology, oceanography, ecology, and geology. Effective interdisciplinary collaboration, although challenging due to the diversity of the disciplines involved, is crucial for advancing GFMs. It enhances the models' comprehensiveness and accuracy by integrating a broad spectrum of scientific insights and methodologies. Key strategies to overcome barriers to effective interdisciplinary work include improving communication, integrating diverse models, and developing supportive collaborative platforms.

\subsubsection{Differences in Methodologies} 
Each scientific discipline contributes unique methodologies, tools, and perspectives to ensemble modeling, posing challenges in creating a cohesive modeling effort. For example, integrating data and models from atmospheric scientists with those from hydrologists requires the alignment of different scales, resolutions, and data types commonly employed in each field \cite{knoben2022community}. This integration is crucial for constructing comprehensive models that accurately represent complex interactions across disciplines, thereby enhancing the reliability and utility of the predictions.

\subsubsection{Communication Barriers}
Effective communication across different scientific disciplines is frequently obstructed using discipline-specific jargon and conceptual differences. Terms and concepts often carry distinct meanings in various fields, which can lead to misunderstandings and misinterpretations. Establishing a common language and understanding is crucial for successfully integrating interdisciplinary insights into GFMs. This foundational communication strategy not only enhances collaboration but also significantly improves the accuracy and effectiveness of the models developed.

\subsubsection{Collaborative Platforms}
Establishing effective platforms for collaboration is essential for advancing interdisciplinary research. This requires implementing both technological and organizational solutions. Technologically, shared databases and integrated modeling tools are vital for seamless data exchange and analysis \cite{an2021challenges}. Organizationally, frameworks that promote regular interaction and cooperation among researchers from diverse fields are equally important. Such structures should be designed to support continuous knowledge sharing and collective problem-solving, thereby enhancing the productivity and impact of collaborative efforts.

\subsubsection{Funding and Institutional Supports}
Securing funding for interdisciplinary projects presents significant challenges, as traditional funding sources are often aligned with specific disciplines rather than cross-disciplinary initiatives. Furthermore, robust institutional support is essential to cultivate an environment that not only encourages but also values interdisciplinary collaboration \cite{bergen2019machine}. Institutions must actively develop policies and provide resources that facilitate the integration of diverse scientific perspectives, thereby enhancing the scope and impact of research outcomes.

\subsection{Privacy, Trust and Security Concerns}
As GFMs increasingly rely on large datasets containing potentially sensitive information, addressing concerns related to privacy, trust, and security becomes crucial. These issues are of particular importance when GFMs are deployed in contexts that could significantly impact public policies or individual lives. To advance and secure broader acceptance of GFMs, it is essential to effectively manage these concerns. The geoscience community must adopt stringent data management practices, enhance transparency, and implement robust security measures. Such efforts will not only improve the trustworthiness and ethical standing of GFMs but also facilitate their responsible and beneficial application in society.

\subsubsection{Data Privacy}
GFMs frequently utilize data that may include geographically and personally identifiable information, such as land use patterns, water usage, or individual movements derived from mobile data. It is imperative to ensure that such data is handled in a manner that respects privacy rights. This commitment involves strict adherence to data protection regulations, such as the General Data Protection Regulation (GDPR) in Europe \cite{budryte2021general}. Additionally, the implementation of robust data anonymization and encryption techniques is essential to safeguard privacy. These measures are crucial for maintaining the integrity and trustworthiness of GFMs, thereby enabling their ethical use in research and policy-making.

\subsubsection{Trust in Models}
The reliability of GFMs is crucial for establishing trust among diverse stakeholders, including the public, policymakers, and scientific communities. Trust in these models largely depends on the transparency of the modeling processes, the precision of the outputs, and clear communication regarding the models' capabilities and limitations \cite{chatzimparmpas2020state}. Establishing such trust is vital because misplaced confidence in flawed models can result in poor decision-making, leading to potentially severe repercussions. Therefore, continuous efforts to enhance model accuracy, process transparency, and stakeholder communication are essential to foster a robust foundation of trust.

\subsubsection{Security Concerns}
The growing interconnectivity of modeling tools and data repositories heightens security risks within the geoscience field. Potential cyberattacks could result in data breaches, manipulation of model outputs, or disruptions to essential modeling infrastructure. To secure GFMs effectively, it is crucial to implement comprehensive cybersecurity measures. This security framework should include regular security audits and the development of robust contingency plans to ensure data integrity and continuous model operation. These proactive measures are fundamental to protecting the integrity of geoscience data and the reliability of model outputs.

\subsubsection{Ethical Use of Models}
The ethical implications of model usage and accessibility extend beyond privacy and security concerns. It is crucial that models, especially those influencing significant economic or policy decisions, are developed and deployed responsibly. This responsibility entails careful attention to potential biases in the data or the models themselves, which could result in unfair or harmful outcomes. Ensuring ethical practices involves maintaining transparency in model development processes, rigorously testing for biases, and making these models accessible to a broad range of stakeholders to avoid privileging or disadvantaging any particular group.

\subsubsection{Legal and Regulatory Compliance}
As the application of geoscience models becomes more widespread, adhering to a complex array of international, national, and local regulations poses a significant challenge. These regulations often cover aspects such as data usage, environmental assessments, and the public dissemination of predictive results. For geoscience modeling efforts to maintain their legitimacy and effectiveness, it is crucial to stay well-informed of relevant laws and ensure strict compliance. This includes regularly updating compliance protocols and training for model developers and users to navigate the evolving legal landscape effectively.

\section{Conclusion}
\label{sec:conclusion}

Geoscience Foundation Models (GFMs) represent a paradigm shift in Earth system modeling, leveraging data-centric AI to integrate massive cross-disciplinary datasets and address complexities that traditional models struggle to capture. This review has highlighted GFMs' unique strengths in flexible task specification, diverse input-output capabilities, and multi-modal knowledge representation, while also presenting key construction techniques focusing on transformers, pre-training, and adaptation. The exploration of recent advances in large language, vision, and vision-language models, particularly in remote sensing applications, demonstrates GFMs' broad potential in geoscience. However, significant challenges remain in data integration, model complexity, uncertainty quantification, interdisciplinary collaboration, and addressing privacy, trust, and security concerns. As we move forward, addressing these challenges will be crucial to fully realizing GFMs' potential. The future of GFMs lies in their continued evolution towards more integrated, data-driven modeling approaches, offering a more comprehensive understanding of Earth's complex systems. This progression promises to provide critical insights into pressing global issues such as climate change, natural hazards, and sustainability, ultimately transforming our approach to Earth system science and informing decision-making in the face of global environmental challenges.
% section conclusion (end)

\section*{Acknowledgment}
The authors would like to express their sincere gratitude and acknowledgment to Professor Gregory C. Beroza from Stanford University for his invaluable opinion on this paper.

\bibliographystyle{IEEEtran}
\bibliography{ref.bib}

\begin{IEEEbiographynophoto}{Hao Zhang} received a B.S. degree in Internet of Things Engineering and an M.Eng. degree in Electronic and Information Engineering from Nanchang University, Nanchang, China, in 2017 and 2020, respectively. 
He is now a last-year Ph.D. candidate at Nanjing University of Aeronautics and Astronautics, and he is also a visiting scholar at Nanyang Technological University (NTU), Singapore from Jan. 2024. 
His research interests include deep learning, geoscience, and computer vision. He has published over 20 papers about deep learning and geoscience, including one ESI highly cited paper. 
\end{IEEEbiographynophoto}

\vspace{10pt}

\begin{IEEEbiographynophoto}{Jin-Jian Xu} is now a postdoctoral fellow in the School of Earth Sciences and Engineering, Nanjing University. His research interests include geoscience, deep learning, and generative artificial intelligence. He has published 15 journal papers about deep learning and geoscience, including two ESI highly cited papers. Additionally, he is the recipient of the 75th Geotechnique Anniversary Early Career Award.
\end{IEEEbiographynophoto}

\vspace{10pt}

\begin{IEEEbiographynophoto}{Hong-Wei Cui} received a B.S. degree in computer science from Northwestern Polytechnical University in 2017 and a Ph.D. degree in computer science from Peking University in 2024. He is now a research engineer in Huawei Technologies Co., Ltd, Shanghai. His research interests include deep learning and large-scale computation. 
\end{IEEEbiographynophoto}

\vspace{10pt}

\begin{IEEEbiographynophoto}{Lin Li} received a B.S. degree in Hohai University in 2019. She is now a Ph.D. candidate in the School of Earth Sciences and Engineering, Nanjing University. Her research interests include geoscience and deep learning. 
\end{IEEEbiographynophoto}

\vspace{10pt}

\begin{IEEEbiographynophoto}{Yaowen Yang}
is currently a Professor in the School of Civil and Environmental Engineering (CEE), Nanyang Technological University (NTU), Singapore. He is one of the lead researchers in areas of structural and geotechnical monitoring and small energy harvesting. His research interests include structural health monitoring, aeroelastic and vibration energy harvesting, metamaterials, and uncertainty analysis in structural dynamics. Over the years, Prof Yang has received more than \$12M research fund from funding agencies and industry, and published over 280 journal and conference papers. 
\end{IEEEbiographynophoto}

\vspace{10pt}

\begin{IEEEbiographynophoto}{Chao-Sheng Tang} 
is a Professor at Nanjing University, and President of Youth Committee of Engineering Geology in China. He is a Distinguished Young Scholar by NSFC. His research interests include unsaturated soils, soil-atmosphere interaction, soil desiccation cracking, bio-geotechnique (MICP), optical fiber sensing technique, contaminated soil, unsaturated soil, and fiber reinforced soil. He is the Editorial Board Member of Canadian Geotechnical Journal, Bulletin of Engineering Geology and the Environment, Environmental Geotechnics, Geoenvironmental Disasters.
\end{IEEEbiographynophoto}
        
\vspace{10pt}

\begin{IEEEbiographynophoto}{Niklas Boers}
is a Professor of Earth System Modelling at, Technical University of Munich and Leader of the Future Lab 'Artificial Intelligence in the Anthropocene' at PIK.  He is also associate coordinator of the Horizon 2020 project 'Tipping Points in the Earth System' (TiPES). He has published over 200 papers about Earth system dynamics on \emph{Nature}, \emph{Nature Machine Intelligence}, \emph{Nature Reviews Earth \& Environment}, \emph{Nature Climate Change}, \emph{Nature Communications}, \emph{Geophysical Research Letters}, and \emph{etc}.
\end{IEEEbiographynophoto}

\end{document}